\documentclass{article}

\usepackage{iclr2027_conference,times}
\iclrfinalcopy
\usepackage[utf8]{inputenc}
\usepackage[T1]{fontenc}

\usepackage{amsmath}
\usepackage{amssymb}
\usepackage{booktabs}
\usepackage[table]{xcolor}
\usepackage{graphicx}
\usepackage{subcaption}
\usepackage{microtype}
\usepackage{multirow}
\usepackage{tabularx}
\usepackage{longtable}
\newcolumntype{L}[1]{>{\raggedright\arraybackslash}p{#1}}
\newcolumntype{Y}{>{\raggedright\arraybackslash}X}
\usepackage{enumitem}
\usepackage[most]{tcolorbox}
\usepackage{listings}
\usepackage{wrapfig}
\usepackage{float}
\usepackage[section]{placeins}
\usepackage{needspace}

\usepackage{url}
\usepackage[para]{footmisc}
\usepackage{hyperref}
\usepackage[capitalize,noabbrev]{cleveref}

\definecolor{PrimaryColor}{HTML}{365F7D}
\definecolor{AccentColor}{HTML}{E45749}
\definecolor{SoftBlue}{HTML}{EDF4F8}
\definecolor{SoftRed}{HTML}{FCEDEA}

\lstdefinestyle{prompt}{
  basicstyle=\rmfamily\small\linespread{0.95}\selectfont,
  breaklines=true,
  breakatwhitespace=true,
  columns=fullflexible,
  keepspaces=true,
  emptylines=0,
  showstringspaces=false
}

\newtcblisting[
  use counter*=lstlisting,
  list inside=lol,
  list type=lstlisting
]{promptbox}[3][]{
  enhanced,
  breakable,
  listing only,
  listing engine=listings,
  listing options={
    style=prompt,
    frame=none,
    backgroundcolor={},
    xleftmargin=0pt,
    xrightmargin=0pt,
    aboveskip=0pt,
    belowskip=0pt,
    #1
  },
  title={\lstlistingname~\thelstlisting: #2},
  title after break={\lstlistingname~\thelstlisting: #2 (continued)},
  lines before break=5,
  nameref={#2},
  label={#3},
  list entry={\protect\numberline{\thelstlisting}{\ignorespaces #2}},
  colback=gray!4,
  colframe=gray!50!black,
  colbacktitle=gray!50!black,
  coltitle=white,
  fonttitle=\bfseries\small,
  arc=2mm,
  outer arc=2mm,
  boxrule=0.6pt,
  left=1.2mm,
  right=1.2mm,
  top=0.8mm,
  bottom=0.8mm,
  before skip=0.3\baselineskip,
  after skip=0.3\baselineskip
}

\hypersetup{
  colorlinks=true,
  linkcolor=PrimaryColor,
  citecolor=PrimaryColor,
  urlcolor=AccentColor,
  pdftitle={ExecCritic: Learn to Test, Test to Improve Coding Agents},
  pdfauthor={Leitian Tao, Baolin Peng, Wenlin Yao, Tao Ge, Hao Cheng, Mike Hang Wang, Sharon Li, Jianfeng Gao},
  pdfsubject={Executable verification for coding-agent self-improvement},
  pdfkeywords={coding agents, executable verification, test generation, reinforcement learning, SWE-bench Verified}
}

\newcommand{\method}{\mbox{\textsc{ExecCritic}}}

\newcommand{\pass}{\ensuremath{\mathrm{PASS}}}
\newcommand{\fail}{\ensuremath{\mathrm{FAIL}}}
\title{\method{}: Learn to Test, Test to Improve for Coding agents}

\author{
Leitian Tao$^{1,2}$\thanks{Work done during an internship at Microsoft Research.} \quad
Baolin Peng$^2$ \quad
Haorui Wang$^{1,3}$ \quad
\textbf{Hang Wang}$^2$ \quad
Hao Cheng$^2$ \quad \\
\textbf{Wenlin Yao}$^2$ \quad 
\textbf{Qianhui Wu}$^2$ \quad
\textbf{Tao Ge}$^2$ \quad 
\textbf{Sharon Li}$^1$\thanks{Co-corresponding authors.} \quad
\textbf{Jianfeng Gao}$^2$\footnotemark[2] \\
$^1$University of Wisconsin--Madison \quad
$^2$Microsoft Research \quad
$^3$Georgia Tech
}

\begin{document}

\maketitle

\begin{abstract}
Execution feedback can guide coding agents toward correct repository repairs, but only when the tests capture the behavior requested by the issue. Agent-generated tests can encode incomplete or incorrect behavioral targets; when the same trajectory writes both the patch and the test, their errors can agree and create false confidence. We introduce \method{}, combining a test--verify--revise scaffold with a role-specific reinforcement learning recipe for training agents within it. The scaffold separates test construction from source-code repair: a Test agent independently generates repository-native tests, a fail-closed harness qualifies and freezes them, and a Repair agent revises source code from their execution feedback without changing the tests. Both roles use Qwen-3.5-35B-A3B as the backbone and are trained separately. In \emph{Learn to Test}, the Test agent learns to produce behaviorally valid tests that distinguish correct from incorrect patches. In \emph{Test to Improve}, the Repair agent learns both direct task resolution and feedback-guided revision. On SWE-bench Verified, test quality determines whether feedback helps: holding the base Repair agent fixed, tests from the base Test agent reduce resolved rate from a no-test baseline of $61.2\%$ to $57.3\%$, whereas tests from GPT-5.6-sol raise it to $65.3\%$. Role-specific post-training raises the Qwen Test agent's Base-to-Gold success from $22.2\%$ to $62.2\%$; composing the two post-trained Qwen agents reaches $72.6\%$, an $11.4$-point gain over the original no-test baseline without stronger-model or Oracle feedback at evaluation time. Code is publicly available at \url{https://github.com/MSR-Orchard/execcritic}

\end{abstract}
\begin{figure}[!htbp]
  \centering
  \vspace{-0.3cm}
  \includegraphics[width=0.95\linewidth]{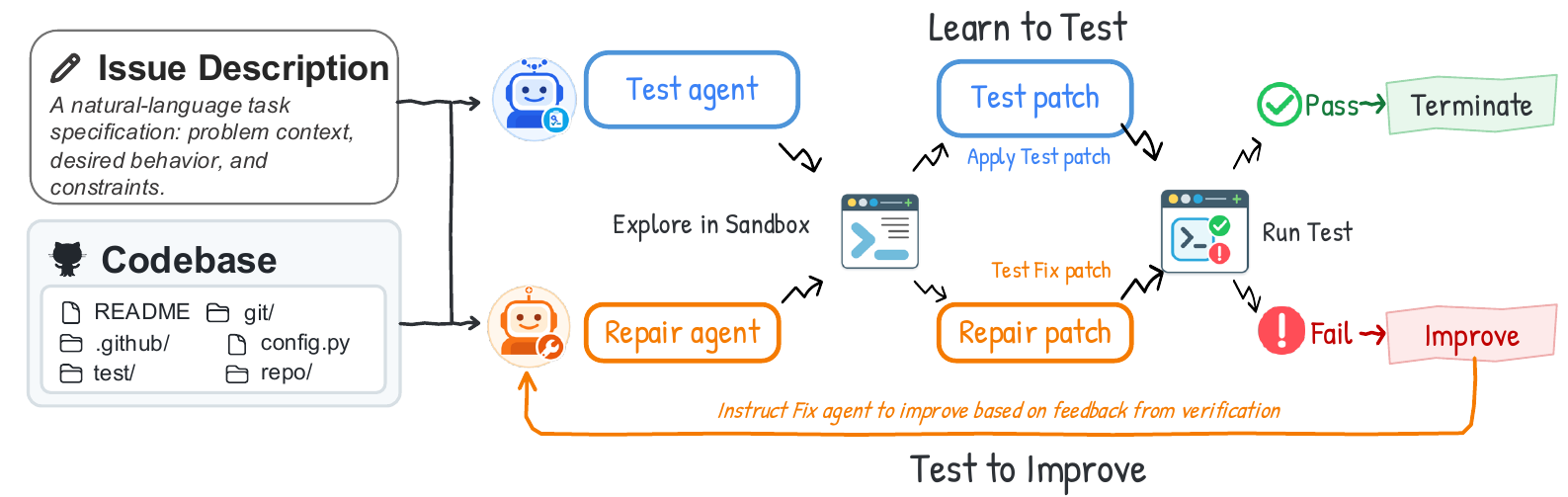}
  \caption{\textbf{\method{} separates test construction from feedback-guided repair.} The Test agent independently constructs a task-specific test, which the harness qualifies and freezes across source-only Repair revisions. Failed checks return execution feedback for revision; a local pass ends revision and requests explicit submission, as summarized by ``Terminate.''}
  \label{fig:framework}
    \vspace{-0.3cm}

\end{figure}
\section{Introduction}
\label{sec:introduction}

Modern AI agents use iterative interaction and feedback to improve performance at inference time~\citep{shinn2023reflexion,chen2023selfdebugging}. This loop is particularly important for repository-level repair: an issue describes intended behavior, but the correct change also depends on implementation details, call sites, and edge cases distributed across a codebase. A failing test can expose missed cases and localize the next revision before submission. Repository-level agents such as SWE-agent~\citep{yang2024sweagent}, OpenHands~\citep{wang2025openhands}, and Mini-SWE-Agent~\citep{lieret2025minisweagent} accordingly interleave exploration, editing, execution, and revision, while prior systems use execution feedback for debugging, candidate selection, bug reproduction, and repair validation~\citep{chen2022codet,ni2023lever,kang2023libro,arora2024masai,xia2025agentless}.

The value of this loop depends on whether its feedback measures the behavior requested by the issue. In repository repair, the official evaluator is hidden, so an agent must construct its own checks and decide when their evidence is sufficient. Agent-written tests may primarily probe runtime behavior rather than assert the requested behavior~\citep{chen2026rethinkingtests}, omit the decisive edge case, or encode the same mistaken interpretation as the source patch. When one trajectory writes both artifacts and decides when to stop, a wrong patch can pass a wrong test and appear validated~\citep{qi2015plausibility,smith2015overfitting,chen2025selfgenerated}. This motivates two requirements: improving the behavioral quality of generated tests and keeping their validation criteria independent of source-patch revision.

Defining this target and repairing source code to satisfy it require different capabilities. Test generation requires inferring a behavioral contract from the issue and repository, expressing it as a discriminative repository-native regression test, and running it with the project's toolchain. Repair requires locating the implementation fault, interpreting possibly imperfect execution feedback, and revising source code without weakening the target. General coding ability does not guarantee either specialization: our off-the-shelf results show that generated feedback can improve or degrade the same Repair agent depending on who generates the test. We therefore train the Test and Repair roles separately. Test training rewards executable tests that distinguish buggy from repaired behavior, whereas Repair training teaches both direct repair and feedback-conditioned revision. Keeping the generated test fixed prevents the Repair agent from weakening the check instead of correcting the source code.

We instantiate this approach in \method{}, a  two-stage framework. In \emph{Learn to Test}, a Test agent explores the repository and produces a repository-native test patch, its exact execution command, and a structured behavior contract. A fail-closed harness qualifies this bundle by validating it and requiring a clean failure on the buggy repository; offline Gold behavior and candidate discrimination provide training signals for learning task-aligned tests. In \emph{Test to Improve}, the harness freezes the qualified Test bundle and executes it against successive source patches from the Repair agent, returning bounded feedback after each attempt. The two stages separate test generation from source-code repair: the Test agent defines the behavioral target, the harness keeps the test fixed, and the Repair agent changes only the source patch. The official evaluator retains final authority over correctness.

Our experiments on SWE-bench Verified show that adding executable feedback is not sufficient: its benefit depends on the quality of the generated tests. Holding the Qwen Repair agent fixed, Qwen-generated tests reduce resolved rate from $61.2\%$ to $57.3\%$, whereas GPT-5.6-generated tests raise it to $65.3\%$ (\cref{sec:off-the-shelf-results}). This contrast motivates learning to construct reliable behavioral targets rather than treating test generation as an off-the-shelf capability. Role-specific post-training improves both components: Test training raises Base-to-Gold success from $22.2\%$ to $62.2\%$, while Repair training raises no-test Round-0 resolution from $61.2\%$ to $68.3\%$. Composing the trained agents reaches $72.6\%$ without stronger-model or Oracle feedback at evaluation---$4.3$ points above the trained Repair agent's no-test result and $11.4$ points above the original baseline (\cref{sec:test-agent-training-results,sec:repair-agent-training-results}). These results support learning both to construct tests and to repair from their feedback. The composed gain reflects the full system, including additional test-generation and revision computation, rather than a compute-matched improvement.

We summarize our contributions below:

\begin{enumerate}
\vspace{-0.2cm}
\item We propose a test--verify--revise scaffold for repository-level repair that separates test construction from source-code revision which qualifies independently generated tests and keeps them fixed while the Repair agent revises source code from execution feedback.

\item We develop a role-specific RL recipe for this scaffold: the Test agent learns to generate behaviorally valid, discriminative tests, while the Repair agent learns direct task resolution and feedback-guided revision.

\item We show that generated-test feedback can help or harm repair depending on test quality, and that composing the trained agents reaches $72.6\%$ on SWE-bench Verified $11.4$ points above the original no-test baseline without stronger-model or oracle feedback at evaluation time.
\end{enumerate}

\section{Motivation: Repair and Validation Are Coupled}
\label{sec:problem}

When a coding agent is asked to fix a bug in a real codebase, it has to produce a patch and decide whether to validate that patch before submission. Since the official evaluator is unavailable, task-specific validation is optional within the agent trajectory. The agent may run checks of its choice, or it may submit without checking the behavior described in the issue. A submitted patch can therefore reach the official evaluator without any local evidence targeted at the requested change.

Running a check introduces a second failure mode.
Suppose a bug occurs only when a function receives an empty list. If the agent overlooks this condition, it may fix the common nonempty case and then write a test covering only that case. The test runs, the patch passes, and the agent concludes that the issue is resolved even though the original bug remains. \emph{The patch and the test agree, but they agree on the same incomplete interpretation of the issue.} Asking the same trajectory to write additional tests does not necessarily resolve this problem: the same blind spot can shape both the solution and the evidence used to validate it. 

\paragraph{A shared trajectory couples repair and stopping evidence.} Formally, let an instance $x=(d,R_B)$ contain an issue description $d$ and a buggy repository checkout $R_B$, called \emph{Base}. 

Before submission, let $\mathcal{E}$ collect the checks and outputs used to assess a Repair patch $p$, with $\mathcal{E}=\varnothing$ when the agent submits without running a check. In a conventional agent-controlled loop, the Repair patch and any validation evidence arise from the same policy and trajectory:
\begin{equation}
  \underbrace{(p,\,\mathcal{E})}_{\text{patch and validation evidence}}
  \sim \pi_\omega(\cdot\mid x).
  \label{eq:coupling}
\end{equation}
This formulation captures both limitations. When $\mathcal{E}=\varnothing$, the patch is submitted without local task-specific validation. When $\mathcal{E}\neq\varnothing$, a shared misinterpretation can shape the patch, the evidence that appears to validate it, and the decision to stop. A check may execute successfully while testing behavior that the issue does not require. Passing such a check establishes that the patch satisfies its assertions. The result does not establish that the patch resolves the task.
The official evaluator $V_x^\star$, which applies hidden fail-to-pass and pass-to-pass tests after submission~\citep{jimenez2024swebench,openai2024swebenchverified}, remains the authority on task correctness, where
$$
V_x^\star(p) \in \{0,1\}
$$

\paragraph{Independent, fixed tests reduce trajectory-level coupling.} We generate each test without access to the candidate Repair trajectory and hold it fixed during source-patch revision. This separation prevents the Repair agent from changing its validation criteria to accommodate a candidate patch. Here, independence refers to the generation context and write permissions, not statistical independence of Test and Repair errors: the two agents can still share a mistaken interpretation of the issue. The Repair agent therefore uses the fixed test's feedback together with the issue and repository behavior when deciding whether to revise or submit a patch.

\method{} operationalizes this separation with a Test agent, a harness, and a Repair agent. The Test agent constructs the check, the harness executes it and keeps it fixed across revisions, and the Repair agent revises only the source patch. Section~\ref{sec:stopping-evidence} describes this two-stage process. 

\section{Methodology: Learn to Test, Test to Improve}
\label{sec:stopping-evidence}

\method{} reduces the coupling between repair and validation by assigning these tasks to separate agents and keeping the generated test fixed during source revision. In practical deployment, the Test agent is intended to provide enough task-specific execution evidence to guide revisions and support the Repair agent's decision to stop and submit. The method has two stages. In \emph{Learn to Test}, the Test agent constructs a focused regression test from the issue, and the harness verifies that the test executes and fails cleanly on Base. In \emph{Test to Improve}, the Repair agent uses the output of this fixed test while revising the source patch.

For the task $x=(d,R_B)$ defined in \cref{sec:problem}, both agents receive the issue $d$ and Base repository state $R_B$. They follow separate trajectories and have different write permissions. The Test agent produces a Test bundle $b=(\Delta_b,c_b,\kappa_b)$: a repository-native Test patch $\Delta_b$, an exact execution command $c_b$, and a JSON behavior contract $\kappa_b$. A Test submission is an attempt to provide this bundle. One bundle targets one coherent issue behavior and may contain multiple test nodes and assertions; a test node is an individually selectable test function, method, or parameterized case. The Repair agent produces a source-only Repair patch $p$. For a valid execution, the harness returns validation evidence $\mathcal{E}(b,p)$ consisting of the test result and execution output. The official evaluator $V_x^\star(p)$ determines whether a submitted Repair patch resolves the task.

\subsection{Learn to Test: Constructing Executable Regression Tests}
\label{sec:test-generation-loop}
\label{sec:harness-gated-execution}

\begin{figure}[t]
  \centering
  \includegraphics[width=\linewidth]{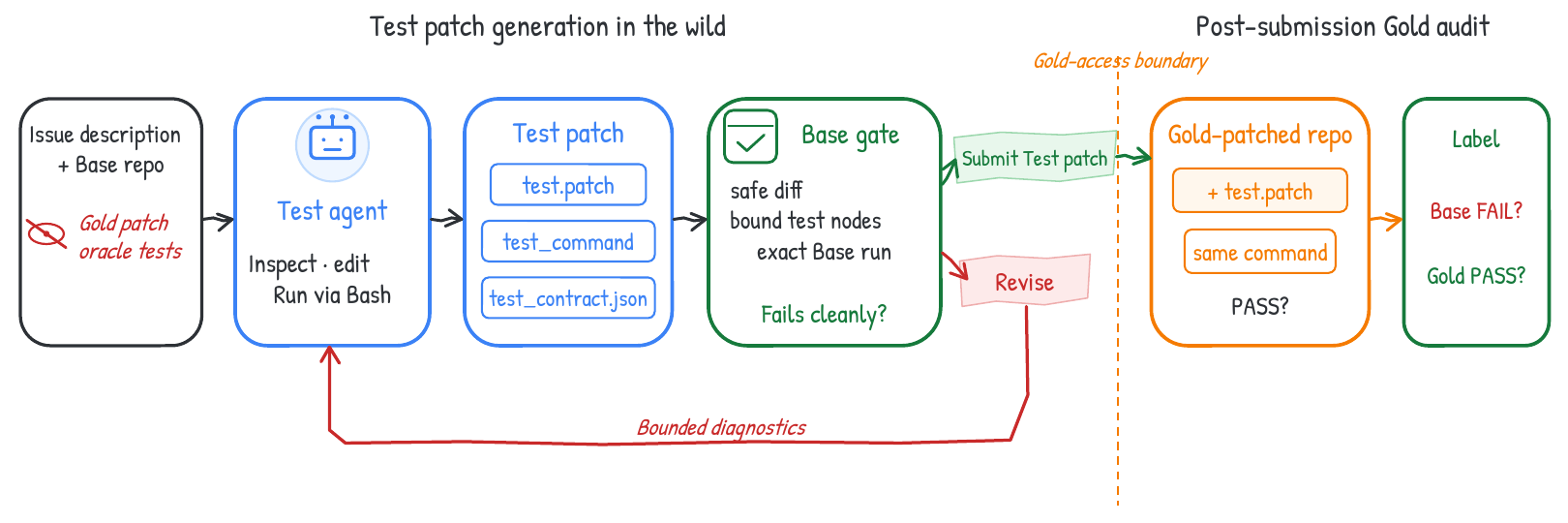}
  \caption{\textbf{Test-submission qualification and offline audit in Learn to Test.} Left: the Test agent uses only the issue and Base checkout to submit a test diff, an exact execution command, and a JSON behavior contract. The harness validates the bundle and requires a clean Base failure, returning bounded diagnostics for unsuccessful attempts. Right: an isolated offline audit runs the unchanged bundle on Gold to measure Base-to-Gold success. Gold outcomes neither enter the generation trajectory nor determine admission to Repair.}
  \label{fig:verifier-construction-harness}
\end{figure}

Learn to Test constructs an executable regression-test bundle before feedback-guided repair begins. Given the public task context $x=(d,R_B)$, the Test policy generates a Test bundle $b$:
\begin{equation}
  b\sim\pi^\mathrm{T}_{\phi}(\cdot\mid x).
  \label{eq:learn-to-test-policy}
\end{equation}
The bundle contains the three artifacts listed in \cref{tab:verifier-artifacts}; \cref{app:evidence-artifacts} documents the persisted interface in detail. The harness validates these items and runs the declared test nodes on a clean copy of $R_B$. Within one generation episode, the Test agent may revise a submission that is invalid or passes on $R_B$ and may submit at most five attempts. The first valid submission that fails cleanly on $R_B$ is retained. If no attempt produces a clean Base failure within the generation budget, the harness marks the instance as a \emph{Base-gate failure} and does not launch feedback-guided Repair. Instead, it retains the Round-0 source patch and sends that patch to the official evaluator. Such instances remain in the all-task resolved-rate denominator. \Cref{fig:verifier-construction-harness} summarizes the complete qualification and offline-audit workflow.

A clean Base failure shows that $R_B$ violates the test expectation, but the expectation may still be incorrect or unrelated to the issue. During offline training and evaluation, we therefore also run $b$ on the Gold repository state $R_G$, obtained by applying the reference Repair patch to $R_B$. Let $B_x(b)=1$ denote a clean failure on $R_B$ and $G_x(b)=1$ denote passage on $R_G$. Base-to-Gold success is
\begin{equation}
  Q_x(b)=B_x(b)G_x(b).
  \label{eq:b2g-gate}
\end{equation}
This outcome shows that the test rejects the Base behavior and accepts the behavior produced by the reference repair. A validation or execution error sets the corresponding indicator to zero. Offline evaluation also runs $b$ on labeled candidate Repair patches to measure whether it passes correct patches and fails incorrect ones. These results are used only for training rewards and analysis and remain outside the Test-agent trajectory. At deployment and in downstream generated-test repair, the harness admits $b$ solely when $B_x(b)=1$. Neither the Gold patch nor its execution outcome is used to filter Test samples or select inputs for the Repair stage.

\begin{figure}[!t]
  \centering
  \includegraphics[width=\linewidth]{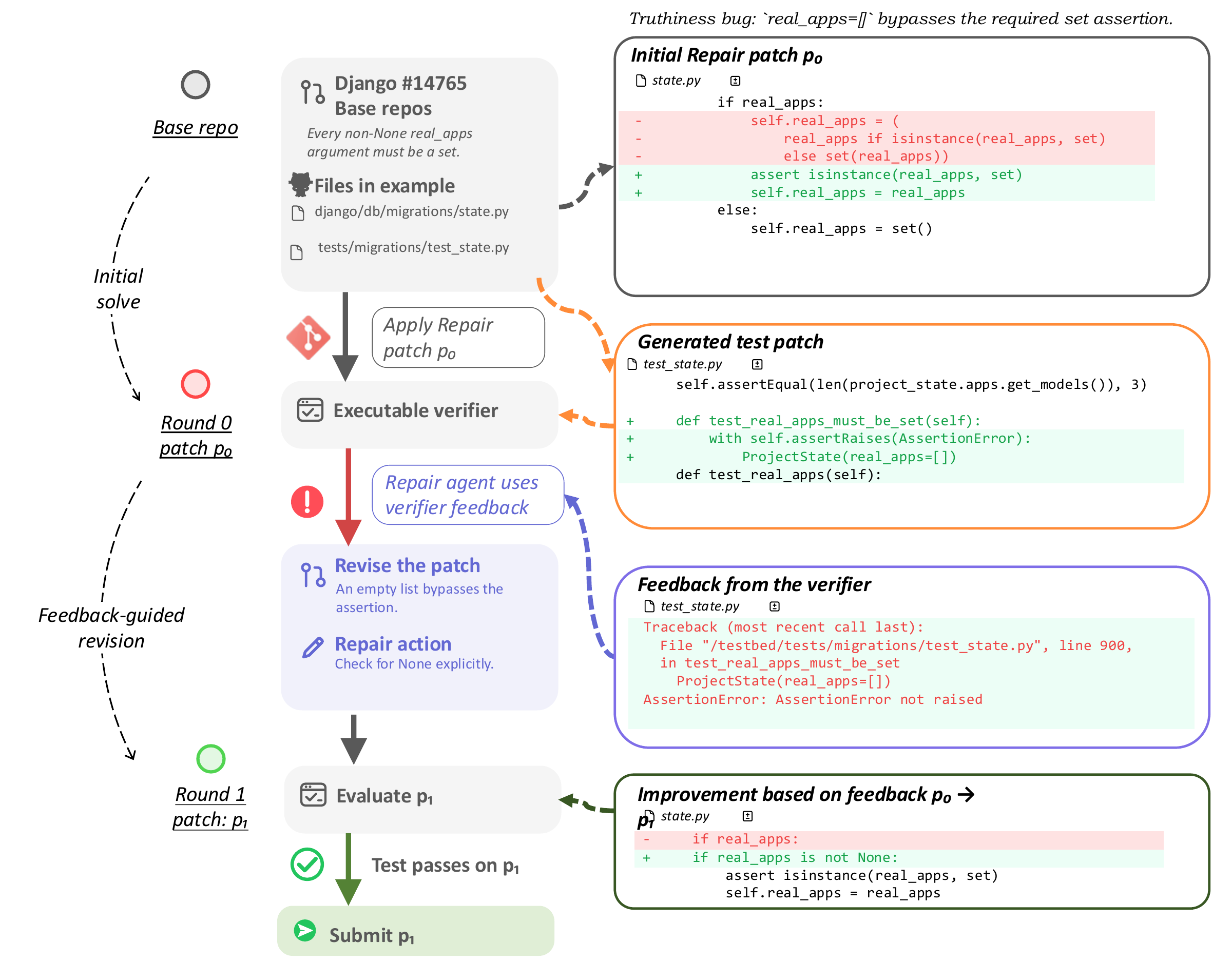}
  \caption{\textbf{A fixed regression test exposes an empty-list bug in Django \#14765.} The initial patch $p_0$ checks the type of \texttt{real\_apps} only when it is truthy, so an empty list bypasses the required assertion. The generated test expects an \texttt{AssertionError} and fails on $p_0$. Changing the guard to \texttt{if real\_apps is not None} produces $p_1$, which passes the same test and is submitted for official evaluation. ``Executable verifier'' denotes the harness.}
  \label{fig:feedback-guided-repair}
\end{figure}

Qualification is necessarily limited by the current harness: artifact validity, declared-node binding, and a clean Base failure do not by themselves prove that the test is semantically aligned with the issue. To evaluate future changes to these validation rules using matched execution records, \cref{app:harness-evolution} specifies an evidence-gated harness-evolution mechanism. This mechanism governs maintenance of the harness rather than the evaluated Test-to-Improve runtime, and no autonomous harness update occurs during an agent trajectory.

\FloatBarrier
\subsection{Test to Improve: Revising Repairs with Fixed Test Feedback}
\label{sec:test-to-improve-loop}
\label{sec:official-authority}

The purpose of Test to Improve is to revise a Repair patch through a sequence of test-guided decisions. Let $h_t$ contain the retained conversation, tool calls, and observations available when round $t$ begins, with $h_0=\varnothing$ before the initial solve. Let $s_t$ denote the information available to the Repair policy, and set $T_{\max}=5$ as the maximum number of feedback-guided revisions after the initial patch. For valid test executions, the process is
\begin{equation}
\begin{aligned}
  s_0 &=(x,h_0),
  &p_0 &\sim \pi^\mathrm{R}_{\theta}(\cdot\mid s_0),\\
  \mathcal{E}_t &=(z_t,o_t)=\mathcal{E}(b,p_t),
  &z_t &\in\{\pass,\fail\},\\
  s_{t+1} &=(x,h_{t+1},p_t,\mathcal{E}_t),
  &p_{t+1} &\sim \pi^\mathrm{R}_{\theta}(\cdot\mid s_{t+1}),
  \quad z_t=\fail,\ t<T_{\max}.
\end{aligned}
  \label{eq:test-to-improve-loop}
\end{equation}
Here, $p_0$ is generated from the issue and Base repository. For each $p_t$, the harness resets its isolated execution workspace to $R_B$, applies $p_t$ and the fixed Test bundle $b$, and returns an outcome $z_t$ with bounded execution feedback $o_t$. The retained history $h_{t+1}$ includes the Round-0 context and subsequent interaction through the current check; it is not replaced by the latest feedback alone. When $z_t=\fail$ and the agent chooses to continue rather than submit, this evidence conditions the next revision $p_{t+1}$. Operational errors do not produce a behavioral verdict and cannot establish local acceptance; they are reported separately as bounded diagnostics (\cref{app:harness-semantics}). The Test bundle remains unchanged across all rounds.

When $z_t=\pass$, the controller immediately ends the Repair episode and submits the passing patch without another agent decision. This local pass is a stopping condition, not a claim of official correctness. The agent may also submit a patch with $z_t=\fail$ when the generated test appears incomplete, overly specific, or incorrect. If the verifier has not passed after the fifth revision, the controller ends the episode and force-submits the latest candidate $p_{T_{\max}}$ as the final patch. \Cref{app:harness-semantics} provides the formal execution details, and \cref{app:harness-prompts} records the exact prompts, feedback envelope, and controller messages. \Cref{fig:feedback-guided-repair} shows an example in which the test output identifies an unhandled empty-list case and guides the next source revision; \cref{app:qualitative-analysis} analyzes two additional completed trajectories with fresh official verification. The official evaluator applies $V_x^\star(p_t)$ to the submitted patch and determines task correctness.

\section{Learning to Test and Repair with ExecCritic}
\label{sec:method}

\begin{figure}[!t]
  \centering
  \includegraphics[width=\linewidth]{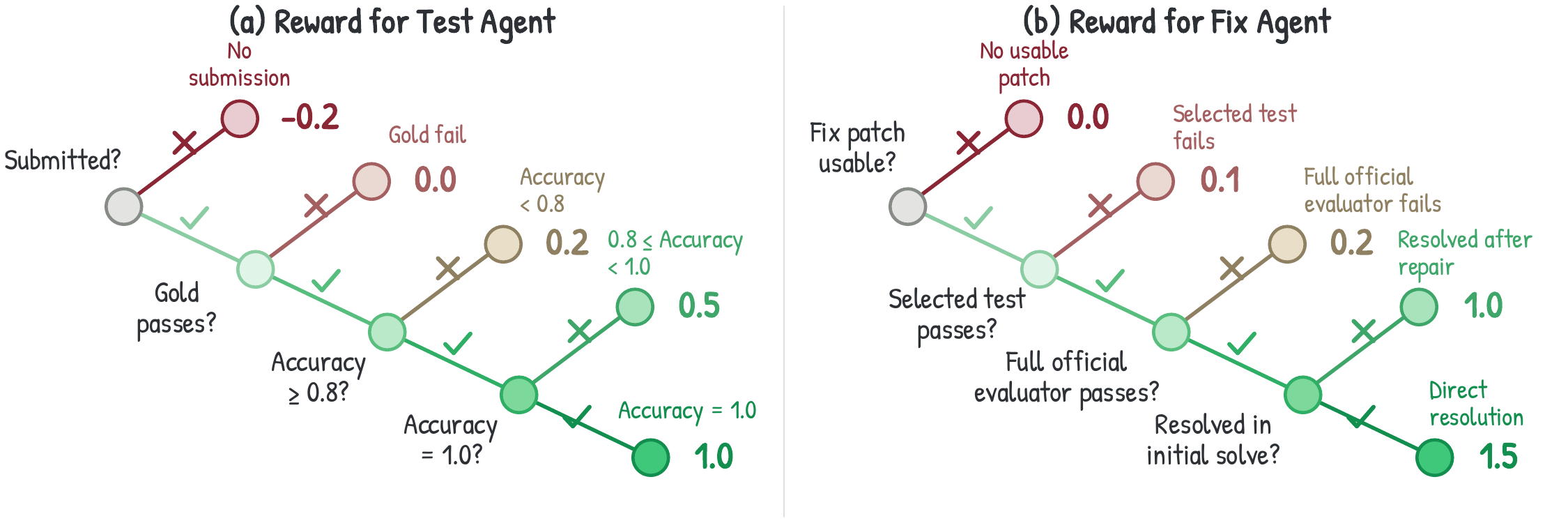}
  \caption{\textbf{Execution-based rewards for Test and Repair training.} (a) For valid submitted bundles, the diagram assumes clean Base failure and assigns reward using Gold passage and balanced accuracy (``Accuracy'') on labeled candidate patches. Values precede the Test trajectory-length adjustment; \cref{sec:learn-to-test-obj} specifies the complete reward and error handling. (b) Repair rewards depend on patch validity, selected-test passage, official success, and whether resolution occurs at Round~0 or after revision. Gold outcomes, candidate labels, and official reward labels remain hidden from the agents; the fixed selected test supplies Repair feedback.}
  \label{fig:reward-trees}
\end{figure}

\method{} develops two complementary capabilities for executable feedback. Test training teaches the model to translate an issue and repository context into a focused, executable regression test. Repair training teaches the model to produce a strong initial fix and to use test outcomes for targeted source revision. We optimize each capability with rewards tied to its observable outcomes, providing direct credit for useful test construction and successful repair. \Cref{fig:reward-trees} summarizes the two training objectives.

\subsection{Learning to Generate Reliable Tests}
\label{sec:learn-to-test-obj}

The Test policy $\pi^\mathrm{T}_{\phi}$ generates a Test bundle $b$ as defined in \cref{sec:test-generation-loop}. We evaluate $b$ using its Base outcome $B_x(b)$, Gold outcome $G_x(b)$, and behavior on labeled candidate Repair patches. Training has two stages. Supervised fine-tuning teaches the submission format and repository workflow. Reinforcement learning then rewards tests that capture the issue behavior and distinguish correct Repair patches from incorrect ones.

\paragraph{Off-policy trajectories teach repository-native test construction.}
Constructing a useful Test patch combines repository exploration, behavioral interpretation, test-infrastructure discovery, and executable artifact generation. We collect high-quality Test trajectories from a fixed stronger model and use them for supervised fine-tuning, providing demonstrations of this complete repository-native workflow. This stage is off-policy because the trajectories are generated by the fixed teacher policy before optimization of the current Test policy. Trajectory admission uses protocol validity and clean Base failure only. Gold outcomes do not filter the SFT trajectories and are not included in the student input.

\paragraph{On-policy RL rewards behavioral validity and discrimination.}
After supervised fine-tuning, we use GRPO with the offline executions defined in \cref{sec:test-generation-loop}. For each task $x$, we sample $K$ Test trajectories with bundles $b_1,\ldots,b_K$ from $\pi^\mathrm{T}_{\phi}$. Each $b$ is run on $R_B$, $R_G$, and at most eight deduplicated candidate Repair patches labeled by $V_x^\star$. Correct candidate patches form the positive class: $\mathrm{TPR}$ is the fraction of correct candidates that pass the generated tests, and $\mathrm{TNR}$ is the fraction of incorrect candidates that fail them in valid executions. We compute balanced accuracy as $\mathrm{BA}_x(b)=\tfrac12(\mathrm{TPR}+\mathrm{TNR})$, with recall set to zero when a class is absent. Consequently, a single-class candidate pool has $\mathrm{BA}_x(b)\leq0.5$ under this convention. If any candidate execution encounters an operational error, the entire affected Test trajectory receives zero advantage and is masked from the policy update and group statistics; the error is not counted as a correct rejection or as a behavioral failure. After excluding these anomalous trajectories, the Test reward is
\begin{equation}
  r_x^{\mathrm{T}}(b)=
  \begin{cases}
    -0.2,&\text{completed without a valid submission},\\
    0,&B_x(b)=0\text{ or }G_x(b)=0,\\
    0.2,&Q_x(b)=1,\ \mathrm{BA}_x(b)<0.8,\\
    0.5,&Q_x(b)=1,\ 0.8\le\mathrm{BA}_x(b)<1,\\
    1.0,&Q_x(b)=1,\ \mathrm{BA}_x(b)=1.
  \end{cases}
  \label{eq:test-raw-reward}
\end{equation}
A completed rollout without a valid submission receives reward $-0.2$. A valid submission receives zero reward if it does not fail cleanly on $R_B$ or does not pass on $R_G$. When $Q_x(b)=1$, the reward increases with balanced accuracy on the candidate Repair patches. The $R_G$ outcomes and candidate labels remain outside the Test-agent state. Within each positive reward level, a rollout that exceeds the shortest rollout by more than eight turns receives half of its original reward. The configured format-error branch remained inactive in the reported runs.

\subsection{Learning to Revise Repairs from Execution Feedback}
\label{sec:test-to-improve-obj}

The Repair policy $\pi^\mathrm{R}_{\theta}$ generates the patch sequence $p_0,p_1,\ldots$ defined in \cref{eq:test-to-improve-loop}. In the main training configuration, we select one Oracle fail-to-pass (F2P) test case for each task and keep it fixed across all Repair rounds. The Repair agent observes only its execution feedback; the test source remains hidden. This controlled feedback source provides a consistent target for learning how to interpret execution results and make corrective source revisions. Each rollout allows at most five feedback-guided revision rounds of 40 turns each after $p_0$. If the selected verifier does not pass within this budget, the controller force-submits the latest candidate as the terminal patch, which $V_x^\star$ evaluates using the full official test suite. For valid terminal evaluations, let $u_t,z_t^{\mathrm{train}}\in\{\pass,\fail\}$ denote the official outcome and the selected training-test outcome for the terminal patch $p_t$, with $u_t=\pass$ exactly when $V_x^\star(p_t)=1$. Following the decision order in \cref{fig:reward-trees}b, the Repair reward is
\begin{equation}
  r_x^{\mathrm{R}}(p_t)=
  \begin{cases}
    0,&\text{invalid terminal patch},\\
    0.1,&z_t^{\mathrm{train}}=\fail,\\
    0.2,&z_t^{\mathrm{train}}=\pass,\ u_t=\fail,\\
    1.0,&z_t^{\mathrm{train}}=\pass,\ u_t=\pass,\ t>0,\\
    1.5,&z_t^{\mathrm{train}}=\pass,\ u_t=\pass,\ t=0.
  \end{cases}
  \label{eq:repair-raw-reward}
\end{equation}
A missing or invalid terminal patch, including a forbidden test modification, receives reward $0$. If the terminal test execution is operationally invalid, the affected Repair trajectory instead receives zero advantage, not a behavioral-failure reward (\cref{app:repair-training-config}). For a usable patch with a valid test execution, a valid failure of the selected training test receives $0.1$, regardless of the official outcome. Passing that test without official success receives $0.2$. Passing both the selected test and the full official evaluator receives $1.0$ after revision or $1.5$ at Round~0. The larger Round-0 reward preserves strong direct-repair behavior while the trajectory also learns to benefit from test-guided revision. The same decision order applies to the generated-test training ablation: an officially correct patch rejected by the generated test receives $0.1$, not the official-success bonus. Official resolved-rate reporting nevertheless depends only on $V_x^\star$, independently of this training reward. The gap between $0.2$ and $1.0$ distinguishes satisfying the focused training test from additionally passing the complete official F2P and pass-to-pass suites.

\section{Experiments}
\label{sec:experiments}

We evaluate three questions. First, can a fixed Test patch raise the resolved rate of an off-the-shelf Repair agent, and how does this effect depend on test quality and benchmark coverage? Second, can post-training increase the Test agent's ability to generate valid behavioral checks? Third, can Repair-agent training strengthen both the initial solution and subsequent feedback-conditioned revision?

\subsection{Experimental Setup}
\label{sec:experimental-setup}

\paragraph{Models and training data.}
We use Qwen-3.5-35B-A3B as the trainable backbone for both roles~\citep{qwen2026qwen35} and GPT-5.6-sol with medium reasoning effort as a stronger-model reference~\citep{openai2026gpt56sol}. We train on SWE-ReBench~\citep{badertdinov2025swerebench} and evaluate on SWE-bench Verified~\citep{jimenez2024swebench,openai2024swebenchverified}. All Test- and Repair-agent training, rollout execution, and reported evaluations are conducted using Orchard, an open-source framework for scalable agentic modeling and reusable sandbox lifecycle management~\citep{peng2026orchard}. Test-agent training retains instances with at least six candidate patches. Repair-agent training additionally requires both correct and incorrect candidate patches, ensuring that the selected examples contain informative revision outcomes.

\paragraph{Test-agent setting.}
We first perform supervised fine-tuning on $5{,}000$ chain-of-thought trajectories generated by DeepSeek-V4-Flash-0731~\citep{deepseek2026v4flash0731} on SWE-ReBench, followed by 200 GRPO steps on SWE-ReBench. Following DAPO-style dynamic sampling~\citep{yu2025dapo}, each update retains 16 issue groups with nonzero within-group reward variance and samples eight rollouts per group, yielding 128 rollouts. During generation, the Test agent may submit at most five attempts. If none produces a clean failure on the Base repository, the loop terminates and the instance is marked as a Base-gate failure sample. Only bundles with a clean Base failure enter downstream feedback-guided repair; otherwise, the Round-0 patch is retained for official scoring. We do not use the Gold patch or Gold outcome to filter this pool. Dynamic sampling keeps issues for which the generated Test bundles receive distinguishable execution rewards, excluding trajectories masked because of candidate-execution errors. Full hyperparameters and executable rewards are detailed in \cref{app:experimental-setup}.

\paragraph{Repair-agent setting.}
We restrict Repair-agent training to SWE-ReBench issues on which the Base model's empirical Round-0 accuracy is below $0.4$, because frequently solved issues provide little revision signal. We train for 100 steps with the same dynamic-sampling configuration: 16 nonzero-variance issue groups and eight rollouts per group. For downstream generated-test repair, an episode is launched only when the Test submission produces a clean Base failure; the Gold patch and Gold outcome do not participate in this selection. Each episode permits at most five feedback-guided repair revision rounds of 40 turns each after the Round-0 patch. If none passes the verifier, the controller force-submits the last candidate as the final patch. This setup emphasizes tasks on which Test-to-Improve can change the terminal outcome while guaranteeing a bounded repair trajectory and a final submission. The rollout and reward configuration is detailed in \cref{app:repair-training-config}.

\paragraph{Evaluation conditions.}
For Test-patch generation, we compare our RL-trained Qwen-3.5-35B-A3B Test agent (RL-35B) with the untrained backbone (Base-35B), its SFT checkpoint, Codex-5.3, the DeepSeek-V4-Flash-0731 teacher, and GPT-5.6-sol. For downstream repair, we compare the Base and Test-to-Improve-trained Qwen Repair agents under five feedback sources: no Test bundle, Base-35B-generated bundles, RL-35B-generated bundles, GPT-5.6-sol-generated bundles, and Oracle fail-to-pass (F2P) tests. The no-test condition measures Round-0 repair capability; Base-35B and RL-35B compare the Test-agent training stages; GPT-5.6-sol provides a stronger-model reference; and Oracle F2P tests provide privileged reference feedback unavailable in the generated-test deployment setting, not a theoretical upper bound. We additionally compare standard Repair training without Test to Improve, Test-to-Improve training without the direct-solve bonus, and the full objective. Test generation is run once per issue and Test source, retaining at most one qualified bundle; Base-to-Gold results describe this single generation run. Resolved rates are means over three Repair runs, which reuse that same bundle for an issue rather than regenerating tests. Every run uses the full benchmark denominator, retaining and officially evaluating the Round-0 patch whenever generation fails the Base gate. The complete evaluation protocol, including fresh official scoring, is specified in \cref{app:evaluation-config}.

\FloatBarrier
\subsection{Main Results}
\label{sec:main-results}

\subsubsection{Off-the-Shelf Agents}
\label{sec:off-the-shelf-results}
\begin{figure}[t]
  \centering
  \setlength{\abovecaptionskip}{4pt}
  \includegraphics[width=\linewidth,trim=0 30bp 0 14bp,clip]{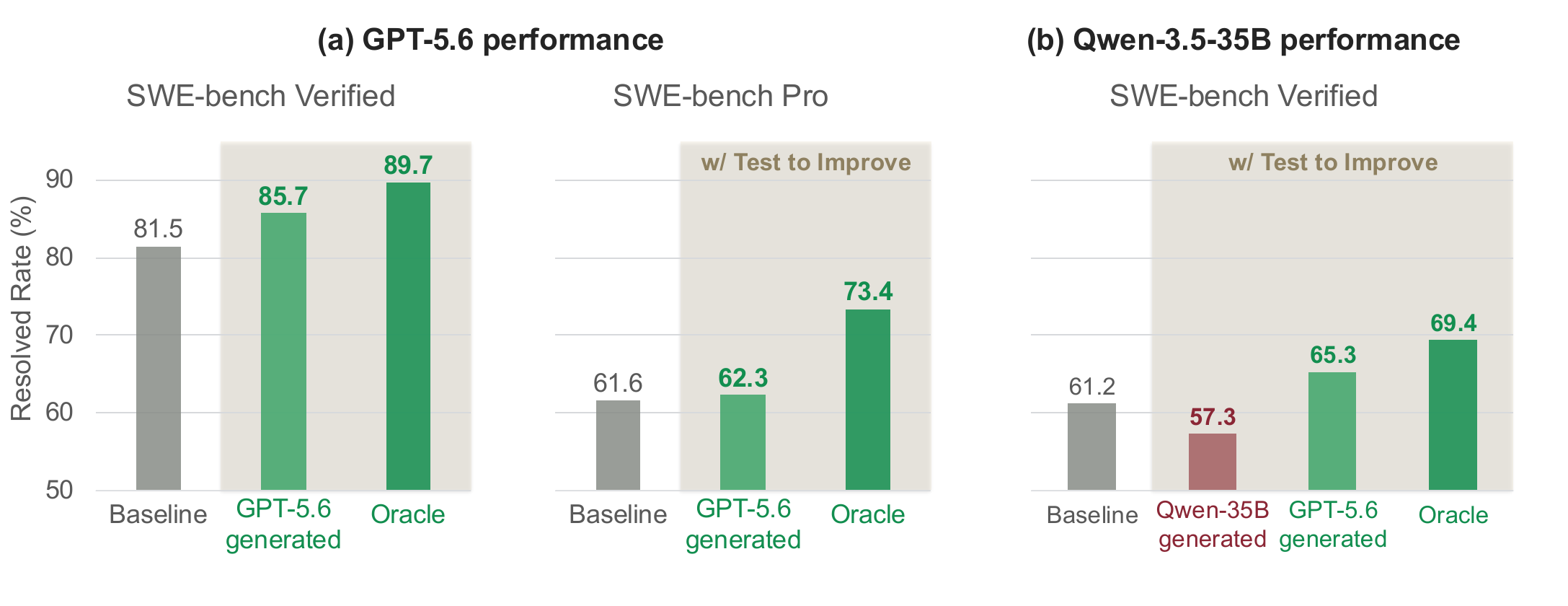}
  \caption{\textbf{Generated-test feedback can help or harm repair.} (a) GPT-5.6 gains more on Verified than Pro. (b) Qwen repair worsens with Qwen tests and improves with GPT-5.6 tests. Baseline: no-test Round~0; Oracle: privileged F2P feedback. Rates (\%) average three all-task Repair runs with fixed bundles; Base-gate failures retain Round~0.}
  \label{fig:test-to-improve-results}
  \vspace{-12pt}
\end{figure}
\paragraph{Generated-test feedback improves frontier-agent repair performance.}
\Cref{fig:test-to-improve-results}a reports results for GPT-5.6. On SWE-bench Verified, generated Test patches raise the resolved rate from $81.5\%$ to $85.7\%$, a gain of $4.2$ points, while Oracle F2P feedback reaches $89.7\%$, a gain of $8.2$ points. On SWE-bench Pro, generated tests improve performance by only $0.7$ points, from $61.6\%$ to $62.3\%$, even though Oracle feedback raises it by $11.8$ points to $73.4\%$. Thus, Pro leaves substantial room for feedback-guided improvement, but feedback from one generated Test bundle recovers much less of that potential. One plausible explanation is that Pro requires broader behavioral coverage: its median and mean numbers of F2P test cases per instance are $3$ and $14.43$, compared with $1$ and $3.03$ for Verified. Although a bundle may contain multiple test nodes and assertions, its focus on one issue behavior may leave other required behaviors uncovered. These aggregate statistics support a coverage-based explanation but do not isolate it from other benchmark differences.

\paragraph{\method{} with open-source agents remains limited by generated test-patch quality.}
\Cref{fig:test-to-improve-results}b holds the Qwen Repair agent fixed and varies only the feedback source. Without a Test patch, the agent resolves $61.2\%$ of tasks. Qwen-generated tests reduce this rate by $3.9$ points to $57.3\%$, whereas GPT-5.6-generated tests improve it by $4.1$ points to $65.3\%$ and Oracle F2P feedback improves it by $8.2$ points to $69.4\%$. This ordering is consistent with the independent Base-to-Gold measurements in \cref{tab:test-agent-b2g}, where GPT-5.6-generated tests are substantially more reliable than tests from the Base Qwen model. With the Repair policy held fixed, these results show that the benefit of execution feedback depends on the Test source. An unreliable test can instead steer revision toward an incomplete or incorrect requirement. This result motivates training the Test agent separately.

\FloatBarrier
\Needspace{15\baselineskip}
\subsubsection{Test-Agent Training Results}
\label{sec:test-agent-training-results}

\begin{wraptable}{r}{0.45\linewidth}
  \vspace{-0.8\baselineskip}
  \centering
  \caption{\textbf{Test-agent Base-to-Gold success (\%) on SWE-bench Verified.} One generation episode per issue and model.}
  \label{tab:test-agent-b2g}
  \footnotesize
  \setlength{\tabcolsep}{3.5pt}
  \renewcommand{\arraystretch}{1.08}
  \begin{tabular}{@{}lr@{}}
    \toprule
    Model & Base$\rightarrow$Gold \\
    \midrule
    Qwen3.5-35B-A3B & 22.2\% \\
    \multicolumn{1}{r}{+ SFT} & 39.6\% \\
    \multicolumn{1}{r}{+ RL} & \textbf{62.2\%} \\
    \midrule
    Codex-5.3 & 61.0\% \\
    DeepSeek-V4-Flash & 73.4\% \\
    GPT-5.6-sol & 87.8\% \\
    \bottomrule
  \end{tabular}
  \vspace{-0.5\baselineskip}
\end{wraptable}

\paragraph{SFT and RL improve Base-to-Gold Test reliability.}
We measure Test reliability by Base-to-Gold success $Q_x$ (\cref{eq:b2g-gate}), which requires a generated test to fail on Base and pass after applying Gold. As shown in \cref{tab:test-agent-b2g}, SFT raises Qwen3.5-35B from $22.2\%$ to $39.6\%$, a gain of $17.4$ points. RL further raises success to $62.2\%$, adding $22.6$ points. The trained Test agent therefore reaches a level comparable to Codex-5.3 at $61.0\%$, while remaining $11.2$ points below DeepSeek-V4-Flash-0731 and $25.6$ points below GPT-5.6-sol. These results separate two effects developed below: SFT establishes the repository workflow and submission behavior, while RL improves the behavioral validity of the policy's own Test submissions.

\begin{figure}[!t]
  \centering
  \includegraphics[width=\linewidth]{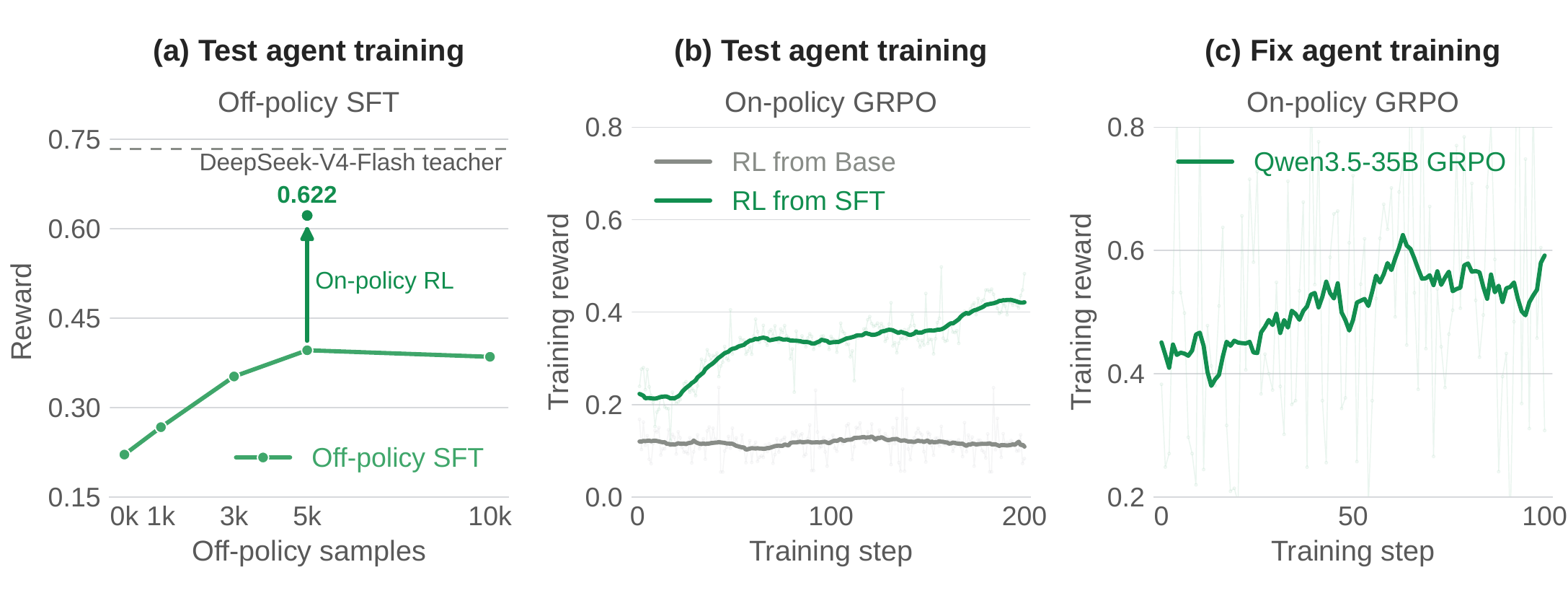}
  \caption{\textbf{Off-policy initialization and on-policy training dynamics.} (a) Test-agent Base-to-Gold success on SWE-bench Verified across SFT dataset sizes; the dashed line is the teacher reference, and the arrow shows RL from the 5K-trajectory checkpoint reaching $0.622$. (b) Test-agent RL reward when initialized from Base or SFT. (c) Repair-agent RL reward.}
  \label{fig:training-triptych}
\end{figure}

\paragraph{On-policy RL adds gains after imitation shows diminishing returns.}
\Cref{fig:training-triptych}a shows that Base-to-Gold success increases from approximately $22\%$ without teacher trajectories to $27\%$ with 1K trajectories, $35\%$ with 3K, and $39\%$ with 5K. Increasing the dataset to 10K produces no further gain in this success metric, while the teacher reference remains near $73\%$. Starting from the 5K SFT checkpoint, on-policy training reaches $62.2\%$ Base-to-Gold success. These values measure behavioral success, not the training reward, which additionally depends on candidate balanced accuracy and trajectory length. The 5K-to-10K comparison suggests diminishing returns from additional imitation at this scale rather than a general saturation of SFT. A likely reason for the subsequent RL gain is that SFT reproduces behaviors represented in fixed teacher trajectories, whereas on-policy execution rewards directly distinguish valid and invalid submissions sampled from the evolving student policy. RL can therefore target errors in the student's own action distribution that additional imitation may not correct.

\paragraph{SFT initialization supplies the behaviors needed for on-policy optimization.}
\Cref{fig:training-triptych}b compares RL initialized from Base and from the SFT checkpoint. In the reported runs, reward remains low and oscillatory from Base but rises steadily from the SFT checkpoint. This pattern is consistent with the structure of the Test reward: a Base policy frequently produces invalid submissions or homogeneous zero-reward outcomes, providing little within-task advantage signal for GRPO. SFT first teaches repository exploration, test construction, and the structured submission protocol, increasing the chance of valid rollouts with differentiated execution outcomes. Off-policy trajectories thus move the policy into a behavioral region where on-policy optimization receives an informative learning signal.

\FloatBarrier
\subsubsection{Repair-Agent Training Results}
\label{sec:repair-agent-training-results}

\begin{table}[!t]
  \caption{\textbf{Resolved rates (\%) on SWE-bench Verified across Repair agents and Test sources.} Both Repair agents use Qwen-3.5-35B-A3B; the trained agent uses the full Test-to-Improve objective. $\Delta$ is the percentage-point change from ``None (Round~0)'' for the same Repair agent. GPT-5.6-sol and Oracle F2P are stronger-model and privileged-feedback references, respectively. Gray marks RL-35B feedback; bold marks the fully trained Test--Repair pair. Values are all-task means over three Repair runs, reusing fixed generated bundles and retaining Round-0 patches when Base qualification fails.}
  \label{tab:test-to-improve-results}
  \centering
  \small
  \setlength{\tabcolsep}{7pt}
  \renewcommand{\arraystretch}{1.15}
  \begin{tabularx}{\linewidth}{@{}Xcccc@{}}
    \toprule
    & \multicolumn{2}{c}{\textbf{Base Repair}} & \multicolumn{2}{c}{\textbf{Trained Repair}} \\
    \cmidrule(lr){2-3}\cmidrule(lr){4-5}
    \textbf{Test Source} & \textbf{Resolved (\%)} & \textbf{$\Delta$ (pp)} & \textbf{Resolved (\%)} & \textbf{$\Delta$ (pp)} \\
    \midrule
    None (Round~0) & 61.2 & --- & 68.3 & --- \\
    GPT-5.6-sol & 65.3 & $+4.1$ & 73.5 & $+5.2$ \\
    Oracle F2P & 69.4 & $+8.2$ & 77.6 & $+9.3$ \\
    \midrule
    Base-35B & 57.3 & $-3.9$ & 64.6 & $-3.7$ \\
    \rowcolor{gray!15}
    RL-35B & 64.1 & $+2.9$ & \textbf{72.6} & $+4.3$ \\
    \bottomrule
  \end{tabularx}
\end{table}
\paragraph{Repair training improves Round-0 performance and increases aggregate feedback gains.}
As shown in \cref{tab:test-to-improve-results}, the no-test Round-0 resolved rate increases from $61.2\%$ for the Base Repair agent to $68.3\%$ for the trained agent, directly demonstrating a $7.1$-point improvement in initial repair capability. With RL-35B feedback, Test to Improve adds $2.9$ points to the Base agent and $4.3$ points to the trained agent, producing final rates of $64.1\%$ and $72.6\%$. Because the two agents begin revision from different Round-0 patches, the difference between these gains is not a matched estimate of feedback-use capability alone. It is consistent with the training objective strengthening both behaviors: the direct-solve reward favors complete initial patches, while fixed-test training exposes the policy to interpreting failure and making corrective source edits. \Cref{fig:training-triptych}c shows an overall increase in Repair training reward, but this trend does not itself establish improved held-out resolution.

\paragraph{Test and Repair training provide complementary system-level gains.}
The comparisons in \cref{tab:test-to-improve-results} vary either the Test or Repair component while holding the other fixed. Replacing Base-35B tests with RL-35B tests raises resolved rate by $6.8$ points for the Base Repair agent and by $8.0$ points for the trained Repair agent. Conversely, Repair training adds $7.3$ points under Base-35B feedback and $8.5$ points under RL-35B feedback. Each trained component therefore improves the system in the presence of either version of the other component, although the displayed averages do not establish a statistically significant interaction between them. Combining the trained Test and Repair agents reaches $72.6\%$ without GPT-5.6 or Oracle F2P feedback during evaluation, $11.4$ points above the original Base Repair agent without feedback. Holding the trained Repair agent fixed, Oracle feedback reaches $77.6\%$. The remaining $5.0$-point gap suggests further gains may be possible through improved test quality and coverage.

\paragraph{Performance gains with limited repair overhead.}
With the trained Test and Repair agents, feedback-guided revision raises resolved rate from $68.3\%$ at Round~0 to $72.6\%$, a gain of $4.3$ percentage points (\cref{tab:test-to-improve-results}). Among the 120 trajectories that entered repair, revision required an average of only 13 additional agent turns. The observed gain therefore accompanies limited additional Repair-agent interaction, rather than routine use of the full revision budget. Test generation is a separate, one-time cost per issue and Test source, with the resulting bundle reused across the three Repair runs. \Cref{app:evaluation-config} reports the revision-turn distribution and evaluation configuration.

\FloatBarrier
\subsection{Additional Analyses and Ablations}

\begin{wraptable}{r}{0.45\linewidth}
  \vspace{-0.8\baselineskip}
  \centering
  \caption{\textbf{Repair-objective ablation on SWE-bench Verified.} Round-0 and final resolved rates (\%) average three runs; final evaluations use fixed RL-35B tests. T2I denotes Test-to-Improve training.}
  \label{tab:repair-training-ablation}
  \footnotesize
  \setlength{\tabcolsep}{3.5pt}
  \renewcommand{\arraystretch}{1.08}
  \begin{tabular}{@{}lrr@{}}
    \toprule
    Training objective & Round 0 & Final \\
    \midrule
    Standard (no T2I) & 68.7 & 70.3 \\
    T2I, no direct bonus & 66.4 & 71.4 \\
    Full T2I & 68.3 & \textbf{72.6} \\
    \bottomrule
  \end{tabular}
  \vspace{-0.5\baselineskip}
\end{wraptable}

\paragraph{The direct-solve bonus balances initial repair and feedback-conditioned revision.}
\Cref{tab:repair-training-ablation} compares standard Repair training, Test-to-Improve training without the direct-solve bonus, and the full objective. Standard training reaches $68.7\%$ at Round~0 and $70.3\%$ after revision. Without the direct-solve bonus, Round-0 performance falls to $66.4\%$ and the final rate reaches $71.4\%$; its larger $5.0$-point within-condition increase partly reflects the lower starting point and does not by itself establish stronger revision capability. The full objective reaches $68.3\%$ at Round~0 and the highest final rate, $72.6\%$. Relative to training without the bonus, it improves Round-0 and final performance by $1.9$ and $1.2$ points, respectively; relative to standard training, it retains comparable Round-0 performance while improving the final rate by $2.3$ points. This pattern is consistent with the direct-solve reward discouraging the policy from deferring an incomplete solution until feedback arrives, while Test-to-Improve training still teaches corrective revision when the initial patch fails.

\begin{wraptable}{r}{0.48\linewidth}
  \vspace{-0.8\baselineskip}
  \centering
  \caption{\textbf{Test-agent SFT teacher comparison.} Single-run Base-to-Gold success (\%) on SWE-bench Verified after SFT and RL. CoT denotes retained teacher reasoning; teacher identity and CoT visibility vary jointly.}
  \label{tab:test-agent-cot}
  \scriptsize
  \setlength{\tabcolsep}{3.2pt}
  \renewcommand{\arraystretch}{1.08}
  \begin{tabular}{@{}lccc@{}}
    \toprule
    SFT teacher & CoT & SFT & $+$ RL \\
    \midrule
    GPT-5.6-sol & No & 35.4 & 36.2 \\
    DeepSeek-V4-Flash & Yes & 39.6 & \textbf{62.2} \\
    \bottomrule
  \end{tabular}
  \vspace{-0.5\baselineskip}
\end{wraptable}

\paragraph{Reasoning-visible SFT trajectories provide a stronger initialization for on-policy learning.}
\Cref{tab:test-agent-cot} compares SFT on GPT-5.6-sol trajectories without chain-of-thought reasoning and DeepSeek-V4-Flash-0731 trajectories that retain it. The two conditions differ by only $4.2$ points after SFT, at $35.4\%$ and $39.6\%$, but the gap expands to $26.0$ points after RL, at $36.2\%$ and $62.2\%$. The larger gap after RL suggests that the two SFT initializations differ more in their support for subsequent policy improvement than in their immediate performance. One possible explanation is that reasoning-visible demonstrations better expose how the teacher interprets repository state, selects actions, and revises its plan after new observations. The result also shows that a stronger standalone teacher need not provide a better initialization for subsequent RL. However, teacher identity, reasoning visibility, and potentially other trajectory properties vary jointly, so this comparison does not isolate a causal effect of CoT.

\begin{wraptable}{r}{0.48\linewidth}
  \vspace{-0.8\baselineskip}
  \centering
  \caption{\textbf{Cross-language Base-to-Gold success (\%).} After Python-only post-training, the Test agent is evaluated once on SWE-bench Verified (Python) and SWE-bench Multilingual subsets (other languages). These are Test-generation, not Repair-resolution, rates.}
  \label{tab:test-agent-cross-language}
  \scriptsize
  \setlength{\tabcolsep}{3.8pt}
  \renewcommand{\arraystretch}{1.08}
  \begin{tabular}{@{}lccccc@{}}
    \toprule
    & Python & Rust & C++ & C & Java \\
    \midrule
    Success & 62.2 & \textbf{66.7} & 58.3 & 26.1 & 2.4 \\
    \bottomrule
  \end{tabular}
  \vspace{-0.5\baselineskip}
\end{wraptable}

\paragraph{Test generation extends beyond Python, with uneven cross-language performance.}
We evaluate the Test agent on non-Python repositories to assess its applicability beyond the language used for post-training. \Cref{tab:test-agent-cross-language} reports Base-to-Gold success on four language subsets of SWE-bench Multilingual, with the $62.2\%$ Python result on SWE-bench Verified included as an in-language reference. The agent reaches $66.7\%$ on Rust and $58.3\%$ on C++, but success is lower on C and Java, at $26.1\%$ and $2.4\%$, respectively. The Rust and C++ results show that the agent can construct repository-native tests that fail on buggy code and pass on reference repairs without additional post-training on those languages. Its applicability is therefore not restricted to Python, although reliability varies substantially across the evaluated task sets and remains limited on C and especially Java.

\Needspace{16\baselineskip}
\begin{wraptable}{r}{0.45\linewidth}
  \vspace{-0.8\baselineskip}
  \centering
  \caption{\textbf{Repair training feedback ablation on SWE-bench Verified.} Round-0 and final resolved rates (\%) average three Repair runs. Both final evaluations use fixed RL-35B-generated tests; the Oracle row uses privileged fail-to-pass feedback only during training.}
  \label{tab:repair-training-feedback-source}
  \footnotesize
  \setlength{\tabcolsep}{4pt}
  \renewcommand{\arraystretch}{1.08}
  \begin{tabular}{@{}lrr@{}}
    \toprule
    Training feedback & Round 0 & Final \\
    \midrule
    Generated tests & 67.6 & 70.9 \\
    Oracle tests & \textbf{68.3} & \textbf{72.6} \\

    \bottomrule
  \end{tabular}
  \vspace{-0.5\baselineskip}
\end{wraptable}

\paragraph{Oracle feedback yields higher Repair performance.}
\Cref{tab:repair-training-feedback-source} compares the fixed feedback source used during Repair training. Generated-test training reaches $67.6\%$ at Round~0 and $70.9\%$ after revision, whereas Oracle-test training reaches $68.3\%$ and $72.6\%$, advantages of $0.7$ and $1.7$ points. This pattern is consistent with feedback quality affecting credit assignment. Generated tests can encode incomplete or incorrect targets, so rewarding patches that satisfy them may reinforce revisions that do not resolve the original task. A selected Oracle F2P test provides a more behaviorally aligned and therefore less noisy revision target, although it still covers only one part of the official specification. Because the terminal reward trains the full trajectory, cleaner feedback can also affect the policy that produces the Round-0 patch, not only its later revisions.

\section{Related Work}
\label{sec:related-work}

\paragraph{Execution-feedback-guided coding agents.}
Long before repository agents, automated testing systems generated inputs from execution feedback or optimized suites for structural coverage, as exemplified by Randoop and EvoSuite~\citep{pacheco2007randoop,fraser2011evosuite}. Automated program repair subsequently exposed a sharper oracle problem: a candidate can be \emph{plausible} because it passes the available suite yet remain semantically incorrect~\citep{qi2015plausibility,smith2015overfitting}. DiffTGen and Opad generate additional cases to reject test-suite-overfitted patches; Patch-Sim compares patch-induced execution behavior; RGT assesses candidates using tests generated from a developer repair; and Poracle checks whether a patch preserves behavior outside the repair region~\citep{xin2017difftgen,yang2017opad,xiong2018patchsim,ye2021rgt,ismayilzada2023poracle}. More recent work clusters patches by generated-test behavior or learns semantic features for filtering overfitting patches~\citep{martinez2024xtestcluster,song2025prism}. At the function level, Self-Debugging and Reflexion revise programs from execution or verbal feedback, while CodeT and LEVER use generated tests or execution results for candidate selection~\citep{chen2023selfdebugging,shinn2023reflexion,chen2022codet,ni2023lever}. At repository scale, constructing task-relevant feedback becomes an agentic problem. LIBRO generates bug-reproducing tests from reports, while SWT-Bench and TDD-Bench Verified operationalize the fail-on-Base/pass-on-Gold criterion for evaluating issue-conditioned tests~\citep{kang2023libro,mundler2024swtbench,ahmed2024tddbench}. Issue2Test, BLAST, and Otter improve reproduction through repository context, search-based testing, and execution-based selection; heterogeneous prompting increases candidate diversity, and SWE-Tester studies post-training open models specifically for issue reproduction~\citep{nashid2026issue2test,kitsios2025blast,ahmed2025otter,ahmed2026heterogeneous,soni2026swetester}. Rather than synthesize every check from scratch, TestPrune localizes and minimizes existing regression tests for both reproduction and patch validation~\citep{chen2026testprune}. Repair systems consume these artifacts in different ways: MASAI and Agentless use reproducers to select or validate patches, CodeR separates reproduction, editing, and verification, and Google's bug-reproduction-test Agent uses generated tests to guide and rank repairs~\citep{arora2024masai,xia2025agentless,chen2024coder,cheng2025agentic}. Repository workflows further differ in whether validation evidence remains fixed. TDFlow freezes supplied or generated tests before source revision, whereas SpecRover, Dynamic Cogeneration, CodeMonkeys, InfCode, Agent-CoEvo, CoHarden, and TDD-Agent update, select, or harden tests as repair proceeds~\citep{han2026tdflow,ruan2025specrover,cheng2026dynamic,ehrlich2025codemonkeys,li2025infcode,li2026agentcoevo,tan2026coharden,yu2026tddagent}. A fixed test supplies a stable revision target; an evolving test can correct weak evidence but also changes the acceptance criterion during the episode. The tradeoff matters because both classical repair studies and recent SWE-bench audits show that tests can admit behaviorally incorrect patches, and agent-written tests may provide observation without strong assertions~\citep{yu2018unsatguided,wang2025solvedissues,yu2025utboost,chen2026rethinkingtests,sun2026swemutation}. SWE-Doctor and EviACT structure or stage execution evidence, while ECLoop and RETRACE place an external critic or gate between a candidate and submission~\citep{guo2026swedoctor,meng2026eviact,xu2026ecloop,li2026retrace}. \method{} uses independently generated tests that remain fixed during repair: it qualifies each Test submission, freezes it during source-only revision, fails closed on operational errors, and leaves final correctness to the official evaluator.

\paragraph{Post-training coding agents for testing and repair.}
Post-training work improves coding agents through successful trajectories, software-evolution data, executable environments, and learned reward signals. CodeLutra iteratively refines code-generation models using preferences between successful and failed code attempts~\citep{tao2025codelutra}. SWE-Fixer separately trains retrieval and editing modules on issue--patch data, while SWE-Dev scales agent trajectories and synthesizes tests for patch evaluation~\citep{xie2025swefixer,wang2025swedev}. SWE-Gym trains repository agents and trajectory verifiers in executable environments, SWE-RL applies reinforcement learning over open software evolution, and Agent-RLVR augments sparse environment rewards with guidance~\citep{pan2025swegym,wei2025swerl,da2025agentrlvr}. Long-context multi-turn RL, SkyRL-Agent, and RepoForge study on-policy repository interaction and scalable SFT--RL pipelines, while LEGO-RL connects native coding-agent harnesses to policy-gradient training and explicitly hardens execution against crashes and reward corruption~\citep{golubev2025longcontext,cao2025skyrlagent,chen2025repoforge,du2026legorl}. More specifically, RLEF trains code models to revise from execution feedback; CURE and UTRL train test generation alongside coding; Code-A1 and Repair-R1 optimize adversarial or test-first code--test interaction; and ReVeal and TaPR train iterative verification or feedback-conditioned repair~\citep{gehring2024rlef,wang2025cure,lee2025utrl,wang2026codea1,hu2025repairr1,jin2025reveal,liu2026tapr}. Related verifier training includes R2E-Gym's hybrid verifiers, CodePRM's execution-enhanced process rewards, and SWE-RM's execution-free reward modeling for test-time selection and RL~\citep{jain2025r2egym,li2025codeprm,shum2025swerm}. Beyond code-specific supervision, HERO combines sparse verifier rewards with dense reward-model scores for reasoning~\citep{tao2026hybrid}, while TRACE derives turn-level rewards from changes in frozen-reference-model state values to address credit assignment in long-horizon tool use~\citep{tao2026trace}. These lines of work improve post-training through execution feedback, learned reward signals, and finer-grained credit assignment. \method{} differs in how it factors the learning problem: the Test policy is trained for Base-to-Gold behavioral change and candidate discrimination, while the Repair policy is trained separately for direct solving and revision from a fixed hidden test. At evaluation time, the two learned roles are composed through a harness that prevents the Repair trajectory from modifying its evidence.

\section{Limitations}
\label{sec:limitations}

The current framework trains the Test and Repair agents as separate policies and composes them only after training. Ideally, both roles would share a single set of model weights, with role conditioning determining whether the model constructs a test or revises a repair. Such a design would simplify deployment by requiring only one checkpoint, allow knowledge to transfer between the two capabilities, and enable end-to-end optimization of the complete test--verify--revise loop. We adopt separate training in this work because the two roles have different action spaces, trajectory structures, and learning signals. Test training rewards valid executable artifacts, Base-to-Gold behavioral change, and candidate discrimination, whereas Repair training rewards source-only solutions and feedback-conditioned revisions that pass the official evaluator. Joint training must balance these heterogeneous signals, assign credit across both trajectories, and prevent degenerate coordination in which the Test and Repair roles adapt to each other rather than to the task specification. Developing a stable shared-weight objective that maintains independent test generation and fixed tests during repair while enabling end-to-end learning is an important direction for future work.

\section{Conclusion}
\label{sec:conclusion}

We introduced \method{}, a framework that separates test generation from source-code repair: the Test agent authors a behavioral check, the harness holds it fixed and executes it, and the Repair agent revises only the source patch from bounded feedback, while the official evaluator retains final authority. Averages over three Repair runs under a shared SWE-bench Verified evaluation protocol show that strong Test patches improve off-the-shelf Repair agents whereas weak ones can hurt, that the Trained Repair condition exceeds Base across the displayed feedback sources, and that both role-specific objectives provide usable optimization signals. These findings support using independently generated, fixed tests to guide repair under the evaluated conditions. Future work will explore multi-check harnesses, stronger Test-agent weights, and improvements in cost and latency.

\bibliographystyle{plainnat}
\bibliography{references}

@inproceedings{jimenez2024swebench,
  title={{SWE}-bench: Can Language Models Resolve Real-World {GitHub} Issues?},
  author={Jimenez, Carlos E. and Yang, John and Wettig, Alexander and Yao, Shunyu and Pei, Kexin and Press, Ofir and Narasimhan, Karthik R.},
  booktitle={International Conference on Learning Representations},
  year={2024}
}

@article{deng2025swebenchpro,
  title={{SWE-Bench Pro}: Can {AI} Agents Solve Long-Horizon Software Engineering Tasks?},
  author={Deng, Xiang and Da, Jeff and Pan, Edwin and He, Yannis Yiming and Ide, Charles and Garg, Kanak and Lauffer, Niklas and Park, Andrew and Pasari, Nitin and Rane, Chetan and Sampath, Karmini and Krishnan, Maya and Kundurthy, Srivatsa and Hendryx, Sean and Wang, Zifan and Bharadwaj, Vijay and Holm, Jeff and Aluri, Raja and Zhang, Chen Bo Calvin and Jacobson, Noah and Liu, Bing and Kenstler, Brad},
  journal={arXiv preprint arXiv:2509.16941},
  year={2025},
  doi={10.48550/arXiv.2509.16941}
}

@article{yang2025swesmith,
  title={{SWE-smith}: Scaling Data for Software Engineering Agents},
  author={Yang, John and Lieret, Kilian and Jimenez, Carlos E. and Wettig, Alexander and Khandpur, Kabir and Zhang, Yanzhe and Hui, Binyuan and Press, Ofir and Schmidt, Ludwig and Yang, Diyi},
  journal={arXiv preprint arXiv:2504.21798},
  year={2025},
  doi={10.48550/arXiv.2504.21798}
}

@misc{openai2024swebenchverified,
  title={Introducing {SWE}-bench Verified},
  author={{OpenAI}},
  year={2024},
  month={August},
  howpublished={\url{https://openai.com/index/introducing-swe-bench-verified/}},
  url={https://openai.com/index/introducing-swe-bench-verified/},
  urldate={2026-08-14},
  note={Blog post}
}

@misc{qwen2026qwen35,
  title={{Qwen3.5}: Towards Native Multimodal Agents},
  author={{Qwen Team}},
  year={2026},
  month={February},
  howpublished={\url{https://qwen.ai/blog?id=qwen3.5}},
  url={https://qwen.ai/blog?id=qwen3.5},
  urldate={2026-08-23},
  note={Blog post}
}

@misc{openai2026gpt56sol,
  title={Previewing {GPT-5.6 Sol}: A Next-Generation Model},
  author={{OpenAI}},
  year={2026},
  month={June},
  howpublished={\url{https://openai.com/index/previewing-gpt-5-6-sol/}},
  url={https://openai.com/index/previewing-gpt-5-6-sol/},
  urldate={2026-08-23},
  note={Blog post}
}

@misc{deepseek2026v4flash0731,
  title={{DeepSeek-V4-Flash-0731}},
  author={{DeepSeek-AI}},
  year={2026},
  url={https://huggingface.co/deepseek-ai/DeepSeek-V4-Flash-0731},
  urldate={2026-09-08},
  note={Model card}
}

@article{peng2026orchard,
  title={{Orchard}: An Open-Source Agentic Modeling Framework},
  author={Peng, Baolin and Yao, Wenlin and Wu, Qianhui and Cheng, Hao and Yu, Xiao and Yang, Rui and Ge, Tao and Sordoni, Alessandro and Yuan, Xingdi and Shen, Yelong and He, Pengcheng and Zhang, Tong and Yu, Zhou and Gao, Jianfeng},
  journal={arXiv preprint arXiv:2605.15040},
  year={2026},
  doi={10.48550/arXiv.2605.15040}
}

@inproceedings{badertdinov2025swerebench,
  title={{SWE-rebench}: An Automated Pipeline for Task Collection and Decontaminated Evaluation of Software Engineering Agents},
  author={Badertdinov, Ibragim and Golubev, Alexander and Nekrashevich, Maksim and Shevtsov, Anton and Karasik, Simon and Andriushchenko, Andrei and Trofimova, Maria and Litvintseva, Daria and Yangel, Boris},
  booktitle={Advances in Neural Information Processing Systems},
  volume={38},
  pages={26420--26466},
  year={2025},
  note={Datasets and Benchmarks Track},
  doi={10.52202/085713-0788}
}

@inproceedings{yu2025dapo,
  title={{DAPO}: An Open-Source {LLM} Reinforcement Learning System at Scale},
  author={Yu, Qiying and Zhang, Zheng and Zhu, Ruofei and Yuan, Yufeng and Zuo, Xiaochen and Yue, Yu and Dai, Weinan and Fan, Tiantian and Liu, Gaohong and Liu, Juncai and Liu, Lingjun and Liu, Xin and Lin, Haibin and Lin, Zhiqi and Ma, Bole and Sheng, Guangming and Tong, Yuxuan and Zhang, Chi and Zhang, Mofan and Zhang, Ru and Zhang, Wang and Zhu, Hang and Zhu, Jinhua and Chen, Jiaze and Chen, Jiangjie and Wang, Chengyi and Yu, Hongli and Song, Yuxuan and Wei, Xiangpeng and Zhou, Hao and Liu, Jingjing and Ma, Wei-Ying and Zhang, Ya-Qin and Yan, Lin and Wu, Yonghui and Wang, Mingxuan},
  booktitle={Advances in Neural Information Processing Systems},
  volume={38},
  pages={125532--125554},
  year={2025},
  doi={10.52202/085713-3775}
}

@inproceedings{yang2024sweagent,
  title={{SWE}-agent: Agent-Computer Interfaces Enable Automated Software Engineering},
  author={Yang, John and Jimenez, Carlos E. and Wettig, Alexander and Lieret, Kilian and Yao, Shunyu and Narasimhan, Karthik and Press, Ofir},
  booktitle={Advances in Neural Information Processing Systems},
  volume={37},
  pages={50528--50652},
  year={2024},
  doi={10.52202/079017-1601}
}

@article{xia2025agentless,
  title={Demystifying {LLM}-Based Software Engineering Agents},
  author={Xia, Chunqiu Steven and Deng, Yinlin and Dunn, Soren and Zhang, Lingming},
  journal={Proceedings of the ACM on Software Engineering},
  year={2025},
  volume={2},
  number={FSE},
  pages={801--824},
  doi={10.1145/3715754}
}

@article{arora2024masai,
  title={{MASAI}: Modular Architecture for Software-engineering {AI} Agents},
  author={Arora, Daman and Sonwane, Atharv and Wadhwa, Nalin and Mehrotra, Abhav and Utpala, Saiteja and Bairi, Ramakrishna and Kanade, Aditya and Natarajan, Nagarajan},
  journal={arXiv preprint arXiv:2406.11638},
  year={2024}
}

@article{chen2024coder,
  title={{CodeR}: Issue Resolving with Multi-Agent and Task Graphs},
  author={Chen, Dong and Lin, Shaoxin and Zeng, Muhan and Zan, Daoguang and Wang, Jian-Gang and Cheshkov, Anton and Sun, Jun and Yu, Hao and Dong, Guoliang and Aliev, Artem and Wang, Jie and Cheng, Xiao and Liang, Guangtai and Ma, Yuchi and Bian, Pan and Xie, Tao and Wang, Qianxiang},
  journal={arXiv preprint arXiv:2406.01304},
  year={2024}
}

@inproceedings{ruan2025specrover,
  title={{SpecRover}: Code Intent Extraction via {LLM}s},
  author={Ruan, Haifeng and Zhang, Yuntong and Roychoudhury, Abhik},
  booktitle={Proceedings of the 47th IEEE/ACM International Conference on Software Engineering},
  year={2025},
  pages={963--974},
  doi={10.1109/ICSE55347.2025.00080}
}

@inproceedings{wang2025openhands,
  title={{OpenHands}: An Open Platform for {AI} Software Developers as Generalist Agents},
  author={Wang, Xingyao and Li, Boxuan and Song, Yufan and Xu, Frank F. and Tang, Xiangru and Zhuge, Mingchen and Pan, Jiayi and Song, Yueqi and Li, Bowen and Singh, Jaskirat and Tran, Hoang H. and Li, Fuqiang and Ma, Ren and Zheng, Mingzhang and Qian, Bill and Shao, Yanjun and Muennighoff, Niklas and Zhang, Yizhe and Hui, Binyuan and Lin, Junyang and Brennan, Robert and Peng, Hao and Ji, Heng and Neubig, Graham},
  booktitle={International Conference on Learning Representations},
  year={2025}
}

@misc{lieret2025minisweagent,
  title={mini-{SWE}-agent},
  author={{SWE-agent Team}},
  year={2025},
  howpublished={\url{https://github.com/SWE-agent/mini-swe-agent}},
  url={https://github.com/SWE-agent/mini-swe-agent},
  urldate={2026-08-14},
  note={Software repository}
}

@inproceedings{pacheco2007randoop,
  title={Feedback-Directed Random Test Generation},
  author={Pacheco, Carlos and Lahiri, Shuvendu K. and Ernst, Michael D. and Ball, Thomas},
  booktitle={International Conference on Software Engineering},
  year={2007},
  pages={75--84},
  doi={10.1109/ICSE.2007.37}
}

@inproceedings{fraser2011evosuite,
  title={{EvoSuite}: Automatic Test Suite Generation for Object-Oriented Software},
  author={Fraser, Gordon and Arcuri, Andrea},
  booktitle={Joint Meeting of the European Software Engineering Conference and the Symposium on the Foundations of Software Engineering},
  year={2011},
  pages={416--419},
  doi={10.1145/2025113.2025179}
}

@inproceedings{shinn2023reflexion,
  title={Reflexion: Language Agents with Verbal Reinforcement Learning},
  author={Shinn, Noah and Cassano, Federico and Gopinath, Ashwin and Narasimhan, Karthik and Yao, Shunyu},
  booktitle={Advances in Neural Information Processing Systems},
  volume={36},
  pages={8634--8652},
  year={2023},
  doi={10.52202/075280-0377}
}

@inproceedings{chen2023selfdebugging,
  title={Teaching Large Language Models to Self-Debug},
  author={Chen, Xinyun and Lin, Maxwell and Sch\"arli, Nathanael and Zhou, Denny},
  booktitle={International Conference on Learning Representations},
  year={2024}
}

@inproceedings{chen2025selfgenerated,
  title={Revisit Self-Debugging with Self-Generated Tests for Code Generation},
  author={Chen, Xiancai and Tao, Zhengwei and Zhang, Kechi and Zhou, Changzhi and Zhang, Xinyu and Gu, Wanli and He, Yuanpeng and Zhang, Mengdi and Cai, Xunliang and Zhao, Haiyan and Jin, Zhi},
  booktitle={Proceedings of the 63rd Annual Meeting of the Association for Computational Linguistics (Volume 1: Long Papers)},
  year={2025},
  pages={18003--18023},
  publisher={Association for Computational Linguistics},
  doi={10.18653/v1/2025.acl-long.881}
}

@inproceedings{qi2015plausibility,
  title={An Analysis of Patch Plausibility and Correctness for Generate-and-Validate Patch Generation Systems},
  author={Qi, Zichao and Long, Fan and Achour, Sara and Rinard, Martin},
  booktitle={Proceedings of the 2015 International Symposium on Software Testing and Analysis},
  year={2015},
  pages={24--36},
  publisher={ACM},
  doi={10.1145/2771783.2771791}
}

@inproceedings{smith2015overfitting,
  title={Is the Cure Worse Than the Disease? Overfitting in Automated Program Repair},
  author={Smith, Edward K. and Barr, Earl T. and Le Goues, Claire and Brun, Yuriy},
  booktitle={Proceedings of the 2015 10th Joint Meeting on Foundations of Software Engineering},
  year={2015},
  pages={532--543},
  publisher={ACM},
  doi={10.1145/2786805.2786825}
}

@inproceedings{xin2017difftgen,
  title={Identifying Test-Suite-Overfitted Patches through Test Case Generation},
  author={Xin, Qi and Reiss, Steven P.},
  booktitle={Proceedings of the 26th ACM SIGSOFT International Symposium on Software Testing and Analysis},
  year={2017},
  pages={226--236},
  publisher={ACM},
  doi={10.1145/3092703.3092718}
}

@inproceedings{yang2017opad,
  title={Better Test Cases for Better Automated Program Repair},
  author={Yang, Jinqiu and Zhikhartsev, Alexey and Liu, Yuefei and Tan, Lin},
  booktitle={Proceedings of the 2017 11th Joint Meeting on Foundations of Software Engineering},
  year={2017},
  pages={831--841},
  publisher={ACM},
  doi={10.1145/3106237.3106274}
}

@inproceedings{xiong2018patchsim,
  title={Identifying Patch Correctness in Test-Based Program Repair},
  author={Xiong, Yingfei and Liu, Xinyuan and Zeng, Muhan and Zhang, Lu and Huang, Gang},
  booktitle={Proceedings of the 40th International Conference on Software Engineering},
  year={2018},
  pages={789--799},
  publisher={ACM},
  doi={10.1145/3180155.3180182}
}

@article{yu2018unsatguided,
  title={Alleviating Patch Overfitting with Automatic Test Generation: A Study of Feasibility and Effectiveness for the {Nopol} Repair System},
  author={Yu, Zhongxing and Martinez, Matias and Danglot, Benjamin and Durieux, Thomas and Monperrus, Martin},
  journal={Empirical Software Engineering},
  year={2019},
  volume={24},
  number={1},
  pages={33--67},
  doi={10.1007/s10664-018-9619-4}
}

@article{ye2021rgt,
  title={Automated Patch Assessment for Program Repair at Scale},
  author={Ye, He and Martinez, Matias and Monperrus, Martin},
  journal={Empirical Software Engineering},
  year={2021},
  volume={26},
  number={2},
  pages={20},
  doi={10.1007/s10664-020-09920-w}
}

@article{ismayilzada2023poracle,
  title={{Poracle}: Testing Patches under Preservation Conditions to Combat the Overfitting Problem of Program Repair},
  author={Ismayilzada, Elkhan and Rahman, Md Mazba Ur and Kim, Dongsun and Yi, Jooyong},
  journal={ACM Transactions on Software Engineering and Methodology},
  year={2024},
  volume={33},
  number={2},
  pages={1--39},
  doi={10.1145/3625293}
}

@article{martinez2024xtestcluster,
  title={Test-Based Patch Clustering for Automatically-Generated Patches Assessment},
  author={Martinez, Matias and Kechagia, Maria and Perera, Anjana and Petke, Justyna and Sarro, Federica and Aleti, Aldeida},
  journal={Empirical Software Engineering},
  year={2024},
  volume={29},
  number={5},
  pages={116},
  doi={10.1007/s10664-024-10503-2}
}

@inproceedings{wang2025solvedissues,
  title={Are ``Solved Issues'' in {SWE-bench} Really Solved Correctly? An Empirical Study},
  author={Wang, You and Pradel, Michael and Liu, Zhongxin},
  booktitle={Proceedings of the 48th IEEE/ACM International Conference on Software Engineering},
  year={2026},
  doi={10.1145/3744916.3764576}
}

@article{song2025prism,
  title={Enhancing {APR} with {PRISM}: A Semantic-Based Approach to Overfitting Patch Detection},
  author={Song, Dowon and Oh, Hakjoo},
  journal={Proceedings of the ACM on Programming Languages},
  year={2025},
  volume={9},
  number={OOPSLA2},
  pages={3342--3370},
  doi={10.1145/3763170}
}

@inproceedings{yu2025utboost,
  title={{UTBoost}: Rigorous Evaluation of Coding Agents on {SWE-Bench}},
  author={Yu, Boxi and Zhu, Yuxuan and He, Pinjia and Kang, Daniel},
  booktitle={Proceedings of the 63rd Annual Meeting of the Association for Computational Linguistics (Volume 1: Long Papers)},
  year={2025},
  pages={3762--3774},
  publisher={Association for Computational Linguistics},
  doi={10.18653/v1/2025.acl-long.189}
}

@inproceedings{sun2026swemutation,
  title={{SWE-Mutation}: Can {LLM}s Generate Reliable Test Suites in Software Engineering?},
  author={Sun, Yuxuan and Zhao, Yuze and Wang, Yufeng and Du, Yao and Ma, Zhiyuan and Wang, Jinbo and Zhang, Mengdi and Zhang, Kai and Huang, Zhenya},
  booktitle={Findings of the Association for Computational Linguistics: ACL 2026},
  year={2026},
  pages={39651--39674},
  publisher={Association for Computational Linguistics},
  doi={10.18653/v1/2026.findings-acl.1976}
}

@inproceedings{kang2023libro,
  title={Large Language Models Are Few-Shot Testers: Exploring {LLM}-Based General Bug Reproduction},
  author={Kang, Sungmin and Yoon, Juyeon and Yoo, Shin},
  booktitle={Proceedings of the 45th International Conference on Software Engineering},
  year={2023},
  pages={2312--2323},
  doi={10.1109/ICSE48619.2023.00194}
}

@inproceedings{mundler2024swtbench,
  title={{SWT-Bench}: Testing and Validating Real-World Bug-Fixes with Code Agents},
  author={M{\"u}ndler, Niels and M{\"u}ller, Mark Niklas and He, Jingxuan and Vechev, Martin},
  booktitle={Advances in Neural Information Processing Systems},
  year={2024},
  volume={37},
  pages={81857--81887}
}

@article{ahmed2024tddbench,
  title={{TDD-Bench Verified}: Can {LLM}s Generate Tests for Issues Before They Get Resolved?},
  author={Ahmed, Toufique and Hirzel, Martin and Pan, Rangeet and Shinnar, Avraham and Sinha, Saurabh},
  journal={arXiv preprint arXiv:2412.02883},
  year={2024}
}

@inproceedings{ahmed2025otter,
  title={{Otter}: Generating Tests from Issues to Validate {SWE} Patches},
  author={Ahmed, Toufique and Ganhotra, Jatin and Pan, Rangeet and Shinnar, Avraham and Sinha, Saurabh and Hirzel, Martin},
  booktitle={Proceedings of the 42nd International Conference on Machine Learning},
  year={2025},
  volume={267},
  pages={752--771},
  publisher={PMLR},
  series={Proceedings of Machine Learning Research}
}

@inproceedings{nashid2026issue2test,
  title={{Issue2Test}: Generating Reproducing Test Cases from Issue Reports},
  author={Nashid, Noor and Bouzenia, Islem and Pradel, Michael and Mesbah, Ali},
  booktitle={Proceedings of the 48th IEEE/ACM International Conference on Software Engineering},
  year={2026}
}

@inproceedings{ahmed2026heterogeneous,
  title={Heterogeneous Prompting and Execution Feedback for {SWE} Issue Test Generation and Selection},
  author={Ahmed, Toufique and Ganhotra, Jatin and Shinnar, Avraham and Hirzel, Martin},
  booktitle={Proceedings of the 48th IEEE/ACM International Conference on Software Engineering},
  year={2026},
  doi={10.1145/3744916.3787837}
}

@article{soni2026swetester,
  title={{SWE-Tester}: Training Open-Source {LLM}s for Issue Reproduction in Real-World Repositories},
  author={Soni, Aditya Bharat and Ghosh, Rajat and Bhargava, Vaishnavi and Chen, Valerie and Dutta, Debojyoti},
  journal={arXiv preprint arXiv:2601.13713},
  year={2026},
  note={Accepted at EMNLP 2026 Industry Track}
}

@article{chen2026testprune,
  title={Can Old Tests Do New Tricks for Resolving {SWE} Issues?},
  author={Chen, Yang and Ahmed, Toufique and Jabbarvand, Reyhaneh and Hirzel, Martin},
  journal={Proceedings of the ACM on Software Engineering},
  year={2026},
  volume={3},
  number={FSE},
  pages={3186--3207},
  doi={10.1145/3808148}
}

@inproceedings{kitsios2025blast,
  title={Automated Generation of Issue-Reproducing Tests by Combining {LLM}s and Search-Based Testing},
  author={Kitsios, Konstantinos and Castelluccio, Marco and Bacchelli, Alberto},
  booktitle={Proceedings of the 40th IEEE/ACM International Conference on Automated Software Engineering},
  year={2025},
  pages={1982--1994},
  publisher={IEEE},
  doi={10.1109/ASE63991.2025.00165}
}

@article{chen2026rethinkingtests,
  title={Rethinking the Value of Agent-Generated Tests for {LLM}-Based Software Engineering Agents},
  author={Chen, Zhi and Sun, Zhensu and Shi, Yuling and Peng, Chao and Gu, Xiaodong and Lo, David and Jiang, Lingxiao},
  journal={arXiv preprint arXiv:2602.07900},
  year={2026}
}

@article{cheng2025agentic,
  title={Agentic Bug Reproduction for Effective Automated Program Repair at {Google}},
  author={Cheng, Runxiang and Tufano, Michele and Cito, J{\"u}rgen and Cambronero, Jos{\'e} and Rondon, Pat and Wei, Renyao and Sun, Aaron and Chandra, Satish},
  journal={arXiv preprint arXiv:2502.01821},
  year={2025}
}

@article{ehrlich2025codemonkeys,
  title={{CodeMonkeys}: Scaling Test-Time Compute for Software Engineering},
  author={Ehrlich, Ryan and Brown, Bradley and Juravsky, Jordan and Clark, Ronald and R{\'e}, Christopher and Mirhoseini, Azalia},
  journal={arXiv preprint arXiv:2501.14723},
  year={2025}
}

@inproceedings{cheng2026dynamic,
  title={Dynamic Cogeneration of Bug Reproduction Test in Agentic Program Repair},
  author={Cheng, Runxiang and Tufano, Michele and Cambronero, Jos{\'e} and Wei, Renyao and Shi, Sherry and Uy, Grant and Rondon, Pat and Ivan{\v c}i{\'c}, Franjo},
  booktitle={Proceedings of the 34th ACM International Conference on the Foundations of Software Engineering},
  series={FSE Companion '26},
  year={2026},
  pages={644--654},
  publisher={ACM},
  doi={10.1145/3803437.3805237}
}

@inproceedings{han2026tdflow,
  title={{TDFlow}: Agentic Workflows for Test Driven Development},
  author={Han, Kevin and Maddikayala, Siddharth and Knappe, Tim and Patel, Om and Liao, Austen and Barati Farimani, Amir},
  booktitle={Proceedings of the 19th Conference of the European Chapter of the Association for Computational Linguistics (Volume 1: Long Papers)},
  year={2026},
  pages={1511--1527},
  publisher={Association for Computational Linguistics},
  doi={10.18653/v1/2026.eacl-long.70}
}

@article{guo2026swedoctor,
  title={{SWE-Doctor}: Guiding Software Engineering Agents with Runtime Diagnosis from Multi-Faceted Bug Reproduction Tests},
  author={Guo, Yaoqi and Liu, Yang and Zhang, Jie M. and Ma, Yun and Lou, Yiling and Chen, Zhenpeng},
  journal={arXiv preprint arXiv:2607.00990},
  year={2026}
}

@article{li2025infcode,
  title={{InfCode}: Adversarial Iterative Refinement of Tests and Patches for Reliable Software Issue Resolution},
  author={Li, Kefan and Wang, Mengfei and Zhang, Hengzhi and Li, Zhichao and Yuan, Yuan and Li, Mu and Gao, Xiang and Sun, Hailong and Hu, Chunming and Lv, Weifeng},
  journal={arXiv preprint arXiv:2511.16004},
  year={2025}
}

@article{li2026agentcoevo,
  title={Beyond Fixed Tests: Repository-Level Issue Resolution as Coevolution of Code and Behavioral Constraints},
  author={Li, Kefan and Yuan, Yuan and Wang, Mengfei and Zheng, Shihao and Wang, Wei and Yang, Ping and Li, Mu and Lv, Weifeng},
  journal={arXiv preprint arXiv:2604.04580},
  year={2026}
}

@article{tan2026coharden,
  title={Beyond Fail-to-Pass: Iterative Hardening of Co-Generated Bug Reproduction Tests and Fixes},
  author={Tan, Yuhao and Yang, Zhibang and Yang, Fangkai and Yao, Yuan and Kang, Yu and Wang, Lu and Zhao, Pu and Zhang, Xin and Ma, Xiaoxing and Lin, Qingwei and Rajmohan, Saravan and Zhang, Dongmei},
  journal={arXiv preprint arXiv:2607.19843},
  year={2026}
}

@inproceedings{chen2022codet,
  title={{CodeT}: Code Generation with Generated Tests},
  author={Chen, Bei and Zhang, Fengji and Nguyen, Anh and Zan, Daoguang and Lin, Zeqi and Lou, Jian-Guang and Chen, Weizhu},
  booktitle={International Conference on Learning Representations},
  year={2023}
}

@inproceedings{ni2023lever,
  title={{LEVER}: Learning to Verify Language-to-Code Generation with Execution},
  author={Ni, Ansong and Iyer, Srini and Radev, Dragomir and Stoyanov, Veselin and Yih, Wen-Tau and Wang, Sida and Lin, Xi Victoria},
  booktitle={Proceedings of the 40th International Conference on Machine Learning},
  year={2023},
  volume={202},
  pages={26106--26128},
  series={Proceedings of Machine Learning Research},
  publisher={PMLR}
}

@article{hu2025repairr1,
  title={{Repair-R1}: Better Test Before Repair},
  author={Hu, Haichuan and Xie, Xiaochen and Zhang, Quanjun},
  journal={arXiv preprint arXiv:2507.22853},
  year={2025}
}

@inproceedings{wang2025cure,
  title={Co-Evolving {LLM} Coder and Unit Tester via Reinforcement Learning},
  author={Wang, Yinjie and Yang, Ling and Tian, Ye and Shen, Ke and Wang, Mengdi},
  booktitle={Advances in Neural Information Processing Systems},
  year={2025},
  volume={38},
  pages={159375--159409},
  doi={10.52202/085713-4809}
}

@inproceedings{lee2025utrl,
  title={Learning to Generate Unit Test via Adversarial Reinforcement Learning},
  author={Lee, Dongjun and Hwang, Changho and Lee, Kimin},
  booktitle={International Conference on Learning Representations},
  year={2026}
}

@inproceedings{jin2025reveal,
  title={{ReVeal}: Self-Evolving Code Agents via Reliable Self-Verification},
  author={Jin, Yiyang and Xu, Kunzhao and Li, Hang and Han, Xueting and Zhou, Yanmin and Li, Cheng and Bai, Jing},
  booktitle={International Conference on Learning Representations},
  year={2026}
}

@article{wang2026codea1,
  title={{Code-A1}: Adversarial Evolving of Code {LLM} and Test {LLM} via Reinforcement Learning},
  author={Wang, Aozhe and Yan, Yuchen and Zhou, Nan and Lu, Zhengxi and Lu, Weiming and Xiao, Jun and Zhuang, Yueting and Shen, Yongliang},
  journal={arXiv preprint arXiv:2603.15611},
  year={2026}
}

@article{lee2026metaharness,
  title={{Meta-Harness}: End-to-End Optimization of Model Harnesses},
  author={Lee, Yoonho and Nair, Roshen and Zhang, Qizheng and Lee, Kangwook and Khattab, Omar and Finn, Chelsea},
  journal={arXiv preprint arXiv:2603.28052},
  year={2026}
}

@article{yu2026tddagent,
  title={{TDD-Agent}: Test-Driven Reasoning for Code Generation},
  author={Yu, Hongyue and Li, Kefan and Li, Jiakun and Chai, Hongzheng and Yuan, Yuan and He, Rui and Wei, Junyi},
  journal={arXiv preprint arXiv:2608.16742},
  year={2026}
}

@inproceedings{ravi2025llmloop,
  title={{LLMLOOP}: Improving {LLM}-Generated Code and Tests Through Automated Iterative Feedback Loops},
  author={Ravi, Ravin and Bradshaw, Dylan and Ruberto, Stefano and Jahangirova, Gunel and Terragni, Valerio},
  booktitle={2025 IEEE International Conference on Software Maintenance and Evolution},
  year={2025},
  pages={930--934},
  publisher={IEEE},
  doi={10.1109/ICSME64153.2025.00109}
}

@inproceedings{gehring2024rlef,
  title={{RLEF}: Grounding Code {LLM}s in Execution Feedback with Reinforcement Learning},
  author={Gehring, Jonas and Zheng, Kunhao and Copet, Jade and Mella, Vegard and Cohen, Taco and Synnaeve, Gabriel},
  booktitle={Proceedings of the 42nd International Conference on Machine Learning},
  series={Proceedings of Machine Learning Research},
  volume={267},
  pages={19034--19055},
  year={2025},
  publisher={PMLR}
}

@article{liu2026tapr,
  title={{TaPR}: Test-Aware Policy Refinement for Feedback-Conditioned Code Generation},
  author={Liu, Aofan and Meng, Jingxiang and Liu, Fangxin and Chen, Yongbiao},
  journal={arXiv preprint arXiv:2608.00494},
  year={2026}
}

@article{da2025agentrlvr,
  title={{Agent-RLVR}: Training Software Engineering Agents via Guidance and Environment Rewards},
  author={Da, Jeff and Wang, Clinton and Deng, Xiang and Ma, Yuntao and Barhate, Nikhil and Hendryx, Sean},
  journal={arXiv preprint arXiv:2506.11425},
  year={2025}
}

@inproceedings{xie2025swefixer,
  title={{SWE}-Fixer: Training Open-Source {LLM}s for Effective and Efficient {GitHub} Issue Resolution},
  author={Xie, Chengxing and Li, Bowen and Gao, Chang and Du, He and Lam, Wai and Zou, Difan and Chen, Kai},
  booktitle={Findings of the Association for Computational Linguistics: ACL 2025},
  year={2025},
  pages={1123--1139},
  publisher={Association for Computational Linguistics},
  doi={10.18653/v1/2025.findings-acl.62}
}

@inproceedings{wang2025swedev,
  title={{SWE}-Dev: Building Software Engineering Agents with Training and Inference Scaling},
  author={Wang, Haoran and Hou, Zhenyu and Wei, Yao and Tang, Jie and Dong, Yuxiao},
  booktitle={Findings of the Association for Computational Linguistics: ACL 2025},
  year={2025},
  pages={3742--3761},
  publisher={Association for Computational Linguistics},
  doi={10.18653/v1/2025.findings-acl.193}
}

@inproceedings{wei2025swerl,
  title={{SWE-RL}: Advancing {LLM} Reasoning via Reinforcement Learning on Open Software Evolution},
  author={Wei, Yuxiang and Duchenne, Olivier and Copet, Jade and Carbonneaux, Quentin and Zhang, Lingming and Fried, Daniel and Synnaeve, Gabriel and Singh, Rishabh and Wang, Sida I.},
  booktitle={Advances in Neural Information Processing Systems},
  volume={38},
  pages={87218--87243},
  year={2025},
  doi={10.52202/085713-2629}
}

@article{golubev2025longcontext,
  title={Training Long-Context, Multi-Turn Software Engineering Agents with Reinforcement Learning},
  author={Golubev, Alexander and Trofimova, Maria and Polezhaev, Sergei and Badertdinov, Ibragim and Nekrashevich, Maksim and Shevtsov, Anton and Karasik, Simon and Abramov, Sergey and Andriushchenko, Andrei and Fisin, Filipp and Skvortsov, Sergei and Yangel, Boris},
  journal={arXiv preprint arXiv:2508.03501},
  year={2025}
}

@article{cao2025skyrlagent,
  title={{SkyRL-Agent}: Efficient {RL} Training for Multi-Turn {LLM} Agent},
  author={Cao, Shiyi and Li, Dacheng and Zhao, Fangzhou and Yuan, Shuo and Hegde, Sumanth R. and Chen, Connor and Ruan, Charlie and Griggs, Tyler and Liu, Shu and Tang, Eric and Liaw, Richard and Moritz, Philipp and Zaharia, Matei and Gonzalez, Joseph E. and Stoica, Ion},
  journal={arXiv preprint arXiv:2511.16108},
  year={2025}
}

@article{chen2025repoforge,
  title={{RepoForge}: Training a {SOTA} Fast-thinking {SWE} Agent with an End-to-End Data Curation Pipeline Synergizing {SFT} and {RL} at Scale},
  author={Chen, Zhilong and Zhao, Chengzong and Chen, Boyuan and Lin, Dayi and Chen, Yihao and Leung, Arthur and Rajbahadur, Gopi Krishnan and Oliva, Gustavo A. and Zhang, Haoxiang and Bhatia, Aaditya and Yong, Chong Chun and Hassan, Ahmed E.},
  journal={arXiv preprint arXiv:2508.01550},
  year={2025}
}

@article{du2026legorl,
  title={{LEGO-RL}: Harness-Native Reinforcement Learning for Coding Agents},
  author={Du, Yiming and Jiang, Yuxin and Yuan, Tao and Dai, Jianbo and Wang, Shaowei and Chen, Jierun and Tao, Chaofan and Yu, Xianzhi and Shang, Lifeng and Wong, Kam-Fai and Li, Xiaohui and Bai, Haoli},
  journal={arXiv preprint arXiv:2608.17393},
  year={2026},
  month={August},
  doi={10.48550/arXiv.2608.17393}
}

@inproceedings{pan2025swegym,
  title={Training Software Engineering Agents and Verifiers with {SWE-Gym}},
  author={Pan, Jiayi and Wang, Xingyao and Neubig, Graham and Jaitly, Navdeep and Ji, Heng and Suhr, Alane and Zhang, Yizhe},
  booktitle={Proceedings of the 42nd International Conference on Machine Learning},
  series={Proceedings of Machine Learning Research},
  volume={267},
  pages={47717--47737},
  year={2025},
  publisher={PMLR}
}

@inproceedings{jain2025r2egym,
  title={{R2E-Gym}: Procedural Environments and Hybrid Verifiers for Scaling Open-Weights {SWE} Agents},
  author={Jain, Naman and Singh, Jaskirat and Shetty, Manish and Zhang, Tianjun and Zheng, Liang and Sen, Koushik and Stoica, Ion},
  booktitle={Conference on Language Modeling},
  year={2025}
}

@article{shum2025swerm,
  title={{SWE-RM}: Execution-Free Feedback for Software Engineering Agents},
  author={Shum, KaShun and Hui, Binyuan and Chen, Jiawei and Zhang, Lei and {X. W.} and Yang, Jiaxi and Huang, Yuzhen and Lin, Junyang and He, Junxian},
  journal={arXiv preprint arXiv:2512.21919},
  year={2025}
}

@article{tao2026trace,
  title={{TRACE}: Turn-level Reward Assignment via Credit Estimation for Long-Horizon Agents},
  author={Tao, Leitian and Peng, Baolin and Yao, Wenlin and Ge, Tao and Cheng, Hao and Wang, Mike Hang and Gao, Jianfeng and Li, Sharon},
  journal={arXiv preprint arXiv:2607.13988},
  year={2026},
}

@article{xu2026ecloop,
  title={Preventing Premature Commitment in Coding Agents with an Evidence-Conditioned Execution Layer},
  author={Xu, Yisen and Li, Chenglin and Wang, Zehao and Yang, Jinqiu and Chen, Tse-Hsun},
  journal={arXiv preprint arXiv:2607.28815},
  year={2026}
}

@article{meng2026eviact,
  title={{EviACT}: An Evidence-to-Action Framework for Agentic Program Repair},
  author={Meng, Qianru and Ren, Zhaochun and Zhang, Xiao and Visser, Joost},
  journal={arXiv preprint arXiv:2605.27238},
  year={2026}
}

@article{li2026retrace,
  title={Independent Patch Verification for Coding Agents with a Bidirectional Reconstruct-and-Verify Framework},
  author={Li, Chenglin and Xu, Yisen and Wang, Zehao and Tan, Shin Hwei and Chen, Tse-Hsun},
  journal={arXiv preprint arXiv:2608.08950},
  year={2026}
}

@inproceedings{li2025codeprm,
  title={{CodePRM}: Execution Feedback-enhanced Process Reward Model for Code Generation},
  author={Li, Qingyao and Dai, Xinyi and Li, Xiangyang and Zhang, Weinan and Wang, Yasheng and Tang, Ruiming and Yu, Yong},
  booktitle={Findings of the Association for Computational Linguistics: ACL 2025},
  year={2025},
  pages={8169--8182},
  publisher={Association for Computational Linguistics},
  doi={10.18653/v1/2025.findings-acl.428}
}

@article{tao2025codelutra,
  title={{CodeLutra}: Boosting {LLM} Code Generation via Preference-Guided Refinement},
  author={Tao, Leitian and Chen, Xiang and Yu, Tong and Mai, Tung and Rossi, Ryan A. and Li, Yixuan and Mitra, Saayan},
  journal={Transactions on Machine Learning Research},
  year={2025},
}

@inproceedings{tao2026hybrid,
  title={Hybrid Reinforcement: When Reward Is Sparse, Better to Be Dense},
  author={Tao, Leitian and Kulikov, Ilia and Saha, Swarnadeep and Wang, Tianlu and Xu, Jing and Li, Sharon and Weston, Jason E. and Yu, Ping},
  booktitle={International Conference on Learning Representations},
  year={2026}
}

\newpage
\appendix
\raggedbottom
\setlength{\parskip}{2pt}
\setlength{\intextsep}{8pt plus 2pt minus 2pt}
\setlength{\textfloatsep}{10pt plus 2pt minus 2pt}
\setlength{\floatsep}{8pt plus 2pt minus 2pt}
\setlength{\LTpre}{4pt}
\setlength{\LTpost}{6pt}
\captionsetup{font=small,skip=4pt}
\section{Experiments and Analysis}
\label{app:experiments}

This appendix provides the reproducibility details behind the training and evaluation results in
\cref{sec:experiments}. We report the Test-agent and Repair-agent configurations separately because
the two roles use different trajectory formats, context and rollout budgets, verification signals,
and reward functions. We then specify the evaluation protocol that composes the independently
trained roles while keeping local generated-test outcomes distinct from official resolution.
The final two subsections document the persisted Test-submission artifacts and analyze two
feedback-guided Repair trajectories that complement the aggregate results.

\subsection{Experimental Setup}
\label{app:experimental-setup}

The Test and Repair policies use Qwen-3.5-35B-A3B and are trained separately. The Test policy is
first supervised on teacher trajectories and is then optimized with GRPO. Its configuration is
reported in \cref{tab:test-agent-hparams}: panels (a)--(b) describe the supervised initialization
and actor optimization, panel (c) specifies repository interaction and verification, panel (d)
defines the executable reward, and panel (e) records trainer and hardware settings.

\begingroup
\footnotesize
\setlength{\tabcolsep}{4pt}
\renewcommand{\arraystretch}{0.98}
\begin{longtable}{@{}L{0.18\linewidth}L{0.29\linewidth}L{\dimexpr0.53\linewidth-4\tabcolsep\relax}@{}}
\caption{\textbf{Test-agent SFT and RL configuration.} The Qwen Test policy is initialized from teacher trajectories and then optimized with execution-based rewards. Panels (a)--(e) specify data, optimization, rollout, reward, and runtime settings.}\label{tab:test-agent-hparams}\\
\toprule
\textbf{Category} & \textbf{Hyperparameter} & \textbf{Value} \\
\midrule
\endfirsthead
\caption[]{\textbf{Test-agent SFT and RL configuration} (continued).}\\
\toprule
\textbf{Category} & \textbf{Hyperparameter} & \textbf{Value} \\
\midrule
\endhead
\bottomrule
\endfoot
\bottomrule
\endlastfoot
\multicolumn{3}{@{}l}{\textbf{(a) Supervised fine-tuning (SFT).}}\\*
\midrule
Model & Student & Qwen-3.5-35B-A3B \\
        & Teacher & DeepSeek-V4-Flash-0731 \\
      \midrule
      Data & Dataset & SWE-ReBench \\
        & Training examples & 5,000 \\
        & Trajectory format & Chain-of-thought Test trajectory \\
      \midrule
      Admission & Submission bundle & Test patch + command + behavior contract \\
        & Required Base outcome & Clean failure \\
        & Gold filtering & Disabled \\
\addlinespace[5pt]
\multicolumn{3}{@{}l}{\textbf{(b) RL data and policy optimization.}}\\*
\midrule
Data & Initialization & Test-agent SFT checkpoint \\
        & Dataset & SWE-ReBench \\
        & Input fields & \texttt{text}, \texttt{patch}, \texttt{metadata} \\
        & Maximum context length & 48,000 tokens \\
        & Maximum response length & 8,192 tokens \\
      \midrule
      Actor Model & Actor & Qwen-3.5-35B-A3B \\
        & Optimizer & Adam \\
        & Adam $\beta_1$ & 0.9 \\
        & Adam $\beta_2$ & 0.98 \\
        & Learning rate & $1\times10^{-6}$ \\
        & Learning-rate schedule & Constant \\
        & Weight decay & 0.1 \\
        & KL coefficient & 0 \\
        & Lower clip ratio & 0.2 \\
        & Upper clip ratio & 0.28 \\
        & Group-mean subtraction & Enabled \\
        & Group-standard-deviation division & Disabled \\
\addlinespace[5pt]
\multicolumn{3}{@{}l}{\textbf{(c) RL rollout and verification.}}\\*
\midrule
Rollout Engine & Backend & SGLang \\
        & GPUs per engine & 1 \\
        & Concurrency & 512 \\
        & Static memory fraction & 0.7 \\
      \midrule
      Sampling & Issues per rollout batch & 16 \\
        & Samples per issue & 8 \\
        & Rollouts per iteration & 128 \\
        & Temperature & 1.0 \\
        & Top-$p$ & 1.0 \\
        & Top-$k$ & $-1$ \\
      \midrule
      Agent Budget & Maximum turns & 60 \\
        & Cost limit & 5 \\
        & Wall-time limit & 900 s \\
      \midrule
      Verifier & Isolation & Independent sandbox \\
        & Install timeout & 1,200 s \\
\addlinespace[5pt]
\multicolumn{3}{@{}l}{\textbf{(d) RL reward.}}\\*
\midrule
Candidate Set & Candidate cap & 8 \\
        & Deduplication & Enabled \\
        & Class balancing & Approximate \\
      \midrule
      Submission Reward & No submission & $-0.2$ \\
        & Valid submission; $Q_x(b)=0$ & 0.0 \\
        & Base-to-Gold; BA$<0.8$ & 0.2 \\
        & Base-to-Gold; $0.8\leq\mathrm{BA}<1$ & 0.5 \\
        & Base-to-Gold; BA$=1$ & 1.0 \\
\addlinespace[5pt]
\multicolumn{3}{@{}l}{\textbf{(e) RL trainer and runtime.}}\\*
\midrule
Trainer & Global batch size & 32 \\
        & Launcher horizon & 200 \\
        & Reported checkpoint & 200 \\
        & Save interval & 10 \\
      \midrule
      Runtime & GPUs per node & 8 \\
        & Nodes & 1 \\
        & Tensor parallelism (TP) & 2 \\
        & Context parallelism (CP) & 4 \\
        & Pipeline parallelism (PP) & 1 \\
        & Expert parallelism (EP) & 4 \\
        & Maximum tokens per GPU & 32,768 \\
\end{longtable}
\endgroup

\paragraph{Test-agent training separates imitation from executable optimization.}
The SFT stage uses 5,000 SWE-ReBench trajectories from DeepSeek-V4-Flash-0731~\citep{deepseek2026v4flash0731}. Admission requires
a clean failure on Base, but does not use Gold behavior as an SFT filter. Starting from this
checkpoint, RL samples eight Test trajectories per issue and executes each submitted bundle in an
independent verifier sandbox. The reward distinguishes invalid submissions, failure to meet the Base-to-Gold criterion,
Base-to-Gold success, and candidate balanced accuracy instead of treating test execution alone as
success. If any candidate execution has an operational error, the entire affected Test trajectory
has zero advantage and is masked from the policy update and group statistics rather than
receiving credit for rejecting that candidate. The 60-turn and 900-second trajectory limits bound repository exploration, while the
turn-slack penalty discourages unnecessarily long successful trajectories. Thus SFT supplies the
initial repository-testing workflow, and RL selects for tests that provide stronger behavioral
evidence under the harness.

\subsection{Repair-Agent Training Configuration}
\label{app:repair-training-config}

Repair training begins from the same model family but optimizes a separate feedback-conditioned
policy. As detailed in \cref{tab:repair-agent-hparams}, each trajectory first receives a 200-turn
Round-0 budget for direct repair and may then use up to five 40-turn revision rounds. A single
deterministically selected Oracle F2P test remains fixed within the rollout, so changes in gate
outcome can be attributed to source-patch revisions rather than to a moving test. The reward gives
the highest value to patches that pass both the selected test and the official evaluator at
Round~0, retains substantial reward when both pass after revision, and assigns only partial
credit to satisfying the focused local test without resolving the task.

\begingroup
\footnotesize
\setlength{\tabcolsep}{4pt}
\renewcommand{\arraystretch}{0.98}
\begin{longtable}{@{}L{0.18\linewidth}L{0.29\linewidth}L{\dimexpr0.53\linewidth-4\tabcolsep\relax}@{}}
\caption{\textbf{Repair-agent RL configuration.} Panels (a)--(d) specify optimization, runtime, feedback-conditioned rollout, and terminal rewards. The main training configuration uses one fixed Oracle fail-to-pass test per rollout, distinct from generated-test feedback at evaluation.}\label{tab:repair-agent-hparams}\\
\toprule
\textbf{Category} & \textbf{Hyperparameter} & \textbf{Value} \\
\midrule
\endfirsthead
\caption[]{\textbf{Repair-agent RL configuration} (continued).}\\
\toprule
\textbf{Category} & \textbf{Hyperparameter} & \textbf{Value} \\
\midrule
\endhead
\bottomrule
\endfoot
\bottomrule
\endlastfoot
\multicolumn{3}{@{}l}{\textbf{(a) Data and actor optimization.}}\\*
\midrule
Data & Dataset & SWE-ReBench \\
        & Problem input field & \texttt{problem\_statement} \\
        & Verifier-only metadata & Enabled \\
        & Maximum context length & 65,536 tokens \\
        & Maximum response length & 6,122 tokens \\
      \midrule
      Actor Model & Actor & Qwen-3.5-35B-A3B \\
        & Optimizer & Adam \\
        & Adam $\beta_1$ & 0.9 \\
        & Adam $\beta_2$ & 0.98 \\
        & Learning rate & $1\times10^{-6}$ \\
        & Learning-rate schedule & Cosine \\
        & Weight decay & 0.1 \\
        & KL coefficient & 0 \\
        & Lower clip ratio & 0.2 \\
        & Upper clip ratio & 0.28 \\
        & Loss reduction & Per-token \\
        & Group-mean subtraction & Enabled \\
        & Group-standard-deviation division & Disabled \\
\addlinespace[5pt]
\multicolumn{3}{@{}l}{\textbf{(b) Trainer and runtime.}}\\*
\midrule
Trainer & Global batch size & 64 \\
        & Launcher horizon & 300 iterations \\
        & Reported checkpoint & 100 \\
        & Save interval & 10 iterations \\
        & Mask aborted rollouts & Enabled \\
        & Zero-variance group filtering & Enabled \\
      \midrule
      Runtime & GPUs per node & 8 \\
        & Nodes & 1 \\
        & Tensor parallelism (TP) & 2 \\
        & Context parallelism (CP) & 4 \\
        & Pipeline parallelism (PP) & 1 \\
        & Expert parallelism (EP) & 4 \\
        & Expert tensor parallelism (ETP) & 1 \\
        & Maximum tokens per GPU & 16,384 \\
\addlinespace[5pt]
\multicolumn{3}{@{}l}{\textbf{(c) Feedback-conditioned rollout.}}\\*
\midrule
Rollout Engine & Backend & SGLang \\
        & GPUs per engine & 2 \\
        & DP-attention data parallelism & 2 \\
        & DP-attention expert parallelism & 2 \\
        & Concurrency & 64 \\
        & Static memory fraction & 0.7 \\
      \midrule
      Sampling & Issues per rollout batch & 16 \\
        & Samples per issue & 8 \\
        & Rollouts per iteration & 128 \\
        & Temperature & 1.0 \\
        & Top-$p$ & 1.0 \\
        & Top-$k$ & $-1$ \\
      \midrule
      Round-0 Budget
        & Turns & 200 \\
      \midrule
      Revision Budget & Rounds & 5 \\
        & Turns per round & 40 \\
      \midrule
      Feedback & Source & Oracle F2P \\
        & Tests per rollout & 1 \\
        & Test selection & Deterministic \\
        & Test persistence & Fixed within each rollout \\
      \midrule
      Verifier & Test timeout & 240 s \\
        & Setup timeout & 600 s \\
        & Feedback cap & 6,000 characters \\
        & Turn-limit action & Force-check current diff \\
\addlinespace[5pt]
\multicolumn{3}{@{}l}{\textbf{(d) Terminal reward.}}\\*
\midrule
Reward & Invalid submission & 0 \\
        & Missing submission & 0 \\
        & Forbidden test modification & 0 \\
        & Usable patch; selected test fails & 0.1 \\
        & Selected test passes; official fails & 0.2 \\
        & Both pass after revision & 1.0 \\
        & Both pass at Round~0 & 1.5 \\
\end{longtable}
\endgroup

\paragraph{Optimization and runtime are separated from the feedback protocol.}
Panels (a)--(b) specify the policy update independently of the later gate interaction. The actor
uses a 65,536-token context and per-token loss reduction, while the trainer forms global batches
of 64 trajectories and masks aborted rollouts. Verifier-only metadata is retained by the training
harness rather than exposed as part of the problem input. Panels (c)--(d) then define how the fixed test, bounded feedback, revision budget, and terminal
reward turn that policy into a feedback-conditioned Repair agent. This separation makes the
source of supervision explicit: model optimization is driven by the terminal task reward, while
the selected test structures the observations available between source-patch revisions.

The focused F2P gate supplies training feedback only; it does not determine official resolution.
The smaller local-only reward prevents passage of the selected training test from being treated as
equivalent to full correctness, while zero reward for forbidden test modification preserves the
source-only Repair boundary used by the runtime controller.

\paragraph{Operationally invalid terminal executions have zero advantage.}
The reward branches in \cref{eq:repair-raw-reward} and \cref{tab:repair-agent-hparams} distinguish usable patches by valid test outcomes. If the final test execution cannot produce a valid outcome because of an environment or verification-infrastructure failure, the affected Repair trajectory $\tau$ receives advantage $A^{\mathrm{R}}(\tau)=0$. This is a final advantage override after any group centering or normalization, not a replacement of the raw reward by zero. The sample therefore contributes no policy-gradient signal and is not assigned the selected-test-failure reward of $0.1$. An unavailable execution result does not establish that the source patch is incorrect. This training rule does not change terminal patch selection: a valid local pass immediately ends repair with the passing patch, and exhaustion of the five revision rounds ends repair with the latest patch.

\subsection{Evaluation Configuration}
\label{app:evaluation-config}

The generated-test evaluation conditions compose the Test and Repair agents without Oracle
feedback, under the protocol in \cref{tab:evaluation-hparams}. Oracle-feedback rows are separate
privileged reference conditions and are not part of this deployment setting. Gold remains hidden
from the Test agent and is used by the harness only for the post-generation audit, never for
qualification or Repair-input selection. For each issue and Test source, one generation episode
retains at most one qualified Test bundle. The same bundle is reused across three Repair rollouts
and all revisions within each rollout; Test generation is not repeated for those rollouts.
Each rollout permits at most five revision rounds of 40 turns each. The bundle's test source remains
hidden from the Repair agent, which receives bounded execution feedback. If generation fails the
Base gate, the corresponding Round-0 patch is retained without feedback-guided revision.
All instances remain in the resolved-rate denominator. A fresh official evaluator scores each
terminal patch, including Round-0 fallbacks, using the complete F2P and P2P suites. Only this
execution determines official resolution; its results are not fed back to the Repair trajectory.
Base-to-Gold percentages describe the single Test-generation run, whereas resolved rates average
the three Repair runs conditional on the fixed generated bundles.

\paragraph{Benchmarks and reported metrics.}
\Cref{tab:evaluation-hparams} specifies the main SWE-bench Verified evaluation. The additional
repair experiments in \cref{fig:test-to-improve-results}a use SWE-bench Pro~\citep{deng2025swebenchpro}
to compare no-test repair, generated-test feedback, and Oracle F2P feedback; their reported metric
is the official resolved rate. The cross-language Test-agent analysis in
\cref{tab:test-agent-cross-language} uses SWE-bench Multilingual~\citep{yang2025swesmith}
for the Rust, C++, C, and Java results. Its Python column reproduces the SWE-bench Verified
Test-agent result as an in-language reference. For this analysis, success is measured by the
Base-to-Gold criterion $Q_x(b)$ in \cref{eq:b2g-gate}: the same generated Test submission must fail
cleanly on Base and pass after applying Gold. These percentages therefore measure Test-generation
reliability, not official Repair-patch resolution, and the reported non-Python columns do not
constitute an aggregate score over all languages in SWE-bench Multilingual.

\begingroup
\footnotesize
\setlength{\tabcolsep}{4pt}
\renewcommand{\arraystretch}{0.98}
\begin{longtable}{@{}L{0.18\linewidth}L{0.29\linewidth}L{\dimexpr0.53\linewidth-4\tabcolsep\relax}@{}}
\caption{\textbf{Generated-test evaluation configuration on SWE-bench Verified.} Panels separate Test generation, feedback-guided Repair, and official scoring. Gold is used only for the offline Test audit, not for admission to Repair; final resolution is determined by the full official evaluator.}\label{tab:evaluation-hparams}\\
\toprule
\textbf{Category} & \textbf{Hyperparameter} & \textbf{Value} \\
\midrule
\endfirsthead
\caption[]{\textbf{Generated-test evaluation configuration on SWE-bench Verified} (continued).}\\
\toprule
\textbf{Category} & \textbf{Hyperparameter} & \textbf{Value} \\
\midrule
\endhead
\bottomrule
\endfoot
\bottomrule
\endlastfoot
\multicolumn{3}{@{}l}{\textbf{(a) Test generation.}}\\*
\midrule
Data & Benchmark & SWE-bench Verified \\
        & Split & \texttt{test} \\
      \midrule
      Agent & Maximum output tokens & 8,192 \\
        & Maximum turns & 60 \\
        & Cost limit & 5 \\
      \midrule
      Parallelism & Workers & 32 \\
        & Endpoint slots & 8 \\
        & Generation runs per issue/source & 1 \\
      \midrule
      Timeouts & Instance timeout & 600 s \\
        & Install timeout & 1,200 s \\
      \midrule
      Harness & Network & Disabled \\
        & Verifier isolation & Independent sandbox \\
        & Gold visibility during generation & Hidden \\
        & Post-generation Gold audit & Enabled \\
      \midrule
      Scoring & Required Base outcome & Clean failure \\
        & Required Gold outcome & Pass \\
        & Base-to-Gold criterion & Both required outcomes \\
\addlinespace[5pt]
\multicolumn{3}{@{}l}{\textbf{(b) Test-to-Improve repair.}}\\*
\midrule
Data & Benchmark & SWE-bench Verified \\
        & Split & \texttt{test} \\
        & Generated Test bundles per issue & At most 1 qualified bundle \\
        & Bundle persistence & Fixed across rollouts and revisions \\
      \midrule
      Sampling & Temperature & 0.6 \\
        & Top-$p$ & Backend default \\
        & Workers & 128 \\
        & Repair rollouts per issue/source & 3 \\
      \midrule
      Revision Budget & Maximum revision rounds & 5 \\
        & Turns per revision round & 40 \\
      \midrule
      Timeouts & Test timeout & 240 s \\
        & Install timeout & 600 s \\
      \midrule
      Harness & Gate sandbox isolation & Independent \\
        & Reset policy & Before grade \\
        & Test source visibility & Hidden from Repair agent \\
        & Returned feedback & Bounded execution output \\
\addlinespace[5pt]
\multicolumn{3}{@{}l}{\textbf{(c) Official scoring.}}\\*
\midrule
Authority & Evaluator & Fresh full official evaluator \\
        & Local-gate authority & Routing only \\
        & Gate-result reporting & Separate \\
      \midrule
      Protocol & F2P suite & All official F2P tests \\
        & P2P suite & All official P2P tests \\
        & Resolved criterion & All required official tests pass \\
        & Base-gate failure & Retain and score Round-0 patch \\
        & Resolved-rate denominator & All benchmark instances \\
\end{longtable}
\endgroup

Local gate outcomes and full official outcomes are recorded separately; only the fresh full
evaluator determines whether a patch is resolved.

\paragraph{Observed revision-turn usage.}
Across the 120 trajectories that entered feedback-guided repair, the number of additional Repair-agent turns averages 13, with a median of 11, a 75th percentile of approximately 17, and a 90th percentile of 34. Of these trajectories, $35.8\%$ use at most five turns and $47.5\%$ use at most ten turns. These statistics measure revision length among trajectories that invoke repair; tasks that do not invoke repair incur no additional revision turns. Test-generation computation is accounted for separately, counting the one-time generation cost once when its bundle is reused across the three Repair rollouts. Turn counts measure agent interaction rather than token usage, wall-clock latency, or monetary cost.

\paragraph{Reproducibility boundary.}
The release includes the launchers, rewards, generated-test submission protocol, repair loops, and evaluation drivers;
datasets, checkpoints, runtime installations, and sandbox credentials remain external.

\subsection{Evidence Artifacts and Executable Contract}
\label{app:evidence-artifacts}
\label{app:verifier-interface}

Beyond aggregate metrics, we record the concrete interface passed from Test generation to the
harness. The three persisted artifacts below bind the proposed behavioral check, its exact
execution target, and its claimed public behavior. The next subsection examines two successful
Repair trajectories built on this interface; \cref{app:formal-specifications} collects the
lower-level harness semantics, maintenance policy, prompts, and controller behavior.

\begin{table}[!htbp]
  \caption{\textbf{Artifacts in a Test submission and their roles in harness validation and execution.} The repository-native diff defines the check, the exact command binds its execution target, and the JSON contract states the intended public behavior.}
  \label{tab:verifier-artifacts}
  \centering
  \small
  \begin{tabularx}{\linewidth}{@{}lXX@{}}
    \toprule
    Artifact & Contents & Harness role \\
    \midrule
    \texttt{test.patch} & Persisted unified diff over repository-native tests & Binds the proposed behavioral check to an auditable code change. \\
    \texttt{test\_command} & Exact test-node selector and invocation & Lets the harness reject commands that target a suite, directory, class, or file. \\
    \texttt{test\_contract.json} & Public entry point, trigger, expected behavior, evidence, excluded interpretations, and declared nodes & States the behavior that the test claims to check and makes its fields and node declarations machine-checkable. \\
    \bottomrule
  \end{tabularx}
\end{table}

\subsection{Qualitative Analysis of Feedback-Guided Repairs}
\label{app:qualitative-analysis}

Each case separates the evidence that exposed the initial patch's failure from the source revision and fresh official verification that followed. An RL-trained Test agent generated the Test patches, and a separately trained Repair agent consumed their execution feedback in the downstream evaluations used for the main results. We selected completed cases whose Test patches exercise the same behavior as the official fail-to-pass tests while using different inputs and assertions. These trajectories illustrate how generated-test feedback can guide successful revisions; they do not establish oracle equivalence or replace the aggregate evaluation.


\par\begingroup
\captionsetup{type=figure}
\begin{tcolorbox}[
  enhanced,
  breakable,
  lines before break=6,
  before skip=5pt,
  after skip=5pt,
  colback=red!5,
  colframe=red!60!black,
  title={\textbf{Case 1: Recursive XOR compilation (Django \#16901)}},
  title after break={\textbf{Case 1: Recursive XOR compilation (Django \#16901)}\space(continued)},
  fonttitle=\small,
  fontupper=\small,
  arc=1.5mm,
  boxrule=0.6pt
]
\textbf{Failure mode.} On databases without native XOR, Django's fallback encoded exactly one true operand. Three or five identical true predicates should match; four should not.\\[2pt]
\textbf{Generated Test patch (40 model calls).} The Test agent added one repository-native test covering all three parity cases. Its exact selector was
\begin{center}
\ttfamily\scriptsize
cd /testbed/tests \&\& python runtests.py\\
queries.tests.Queries1Tests.test\_xor\_with\_multiple\_identical\_q -v 2
\end{center}
The command failed cleanly on Base. The selected test body passed once on Gold, but a later wrapper cleanup command returned code 1 after a \texttt{git checkout} pathspec error.\\[2pt]
\textbf{Initial Repair and execution feedback.} The initial patch computed parity with a modulo expression but rebuilt the fallback node with \texttt{self.connector}. The frozen Test patch exposed recursive XOR compilation:
\begin{center}
\ttfamily\scriptsize
[Previous line repeated 959 more times]\\
RecursionError: maximum recursion depth exceeded in comparison\\
FAILED (errors=1)
\end{center}
\end{tcolorbox}

\vspace{2pt}
\begin{tcolorbox}[
  enhanced,
  breakable,
  lines before break=6,
  before skip=5pt,
  after skip=5pt,
  colback=green!5,
  colframe=green!60!black,
  title={\textbf{Feedback-guided Repair (1 round, 23 turns)}},
  title after break={\textbf{Feedback-guided Repair (1 round, 23 turns)}\space(continued)},
  fonttitle=\small,
  fontupper=\small,
  arc=1.5mm,
  boxrule=0.6pt
]
\textbf{Source revision.} The Repair agent traced the recursion to the reused XOR connector and replaced it with \texttt{AND}:
\begin{center}
\ttfamily\scriptsize
- return self.\_\_class\_\_([rhs], self.connector, self.negated).as\_sql(\\
+ return self.\_\_class\_\_([rhs], AND, self.negated).as\_sql(\\
\hspace*{2em}compiler, connection)
\end{center}
\textbf{Generated-test outcome:} \checkmark{} the same frozen Test patch passed.\\[2pt]
\textbf{Official outcome:} The initial Repair patch failed one official fail-to-pass test and all six pass-to-pass tests. \checkmark{} The terminal patch passed \texttt{test\_filter\_multiple} and all six pass-to-pass tests; the evaluator returned \texttt{resolved=true}. The generated test used repeated predicates, whereas the official test used distinct threshold predicates; both exercised odd parity.
\end{tcolorbox}
\caption{\textbf{A recursion trace guides source repair in Django \#16901.} The initial patch computes parity but reuses the XOR connector, recursively re-entering the same compilation path. Replacing it with \texttt{AND} yields a patch that passes both the frozen generated test and fresh official verification. The archived Gold audit includes a wrapper cleanup error, so this case illustrates successful downstream repair rather than a clean Base-to-Gold audit.}
\label{app:trajectory-django-16901}
\par\endgroup


\par\begingroup
\captionsetup{type=figure}
\begin{tcolorbox}[
  enhanced,
  breakable,
  lines before break=6,
  before skip=5pt,
  after skip=5pt,
  colback=orange!5,
  colframe=orange!70!black,
  title={\textbf{Case 2: Repair targets the wrong parser (Sphinx \#9230)}},
  title after break={\textbf{Case 2: Repair targets the wrong parser (Sphinx \#9230)}\space(continued)},
  fonttitle=\small,
  fontupper=\small,
  arc=1.5mm,
  boxrule=0.6pt
]
\textbf{Failure mode.} Sphinx split a typed parameter field at the comma inside \texttt{dict(str, str)}, separating the parameter name from an incomplete type.\\[2pt]
\textbf{Generated Test patch (40 model calls).} The Test agent added a parser test for \texttt{:param dict(str, str) name:} and required the rendered text to equal \texttt{name (dict(str, str)) -- blah blah}. Its persisted command was
\begin{center}
\ttfamily\scriptsize
python -m pytest\\
tests/test\_domain\_py.py::test\_info\_field\_list\_param\_type\_dict\_str\_str -q
\end{center}
Base failed by rendering \texttt{str) name (dict(str,)}; Gold returned code 0 with one passing test.\\[2pt]
\textbf{Initial Repair and execution feedback.} The initial Repair patch edited the unrelated Napoleon parser, while the failing path remained in \texttt{sphinx/util/docfields.py}. The frozen Test patch localized the observable mismatch:
\begin{center}
\ttfamily\scriptsize
- name (dict(str, str)) -- blah blah\\
+ str) name (dict(str,) -- blah blah\\
1 failed, 7 warnings in 0.21s
\end{center}
\end{tcolorbox}

\vspace{2pt}
\begin{tcolorbox}[
  enhanced,
  breakable,
  lines before break=6,
  before skip=5pt,
  after skip=5pt,
  colback=green!5,
  colframe=green!60!black,
  title={\textbf{Feedback-guided Repair (2 rounds, 75 turns)}},
  title after break={\textbf{Feedback-guided Repair (2 rounds, 75 turns)}\space(continued)},
  fonttitle=\small,
  fontupper=\small,
  arc=1.5mm,
  boxrule=0.6pt
]
\textbf{Source revision.} The Repair agent moved to the typed-field parser and preserved all tokens before the final parameter name:
\begin{center}
\ttfamily\scriptsize
+ parts = fieldarg.split()\\
+ last\_part = parts[-1]\\
+ if all(c.isalnum() or c == '\_' for c in last\_part):\\
+ \hspace*{1em}return ' '.join(parts[:-1]).strip(), last\_part\\
- argtype, argname = fieldarg.split(None, 1)\\
+ argtype, argname = \_split\_type\_and\_name(fieldarg)
\end{center}
\textbf{Generated-test outcome:} \checkmark{} the same frozen Test patch passed.\\[2pt]
\textbf{Official outcome:} The initial Repair patch failed one official fail-to-pass test while all 44 pass-to-pass tests passed. \checkmark{} The terminal patch passed \texttt{test\_info\_field\_list} and all 44 pass-to-pass tests; the evaluator returned \texttt{resolved=true}. The generated test used a parenthesized type, while the official test used a bracketed type through the same parsing path. The terminal patch retained the unrelated Napoleon edit and was therefore successful but nonminimal.
\end{tcolorbox}
\caption{\textbf{A rendering mismatch redirects repair in Sphinx \#9230.} The initial patch changes the unrelated Napoleon parser. Generated-test feedback redirects the Repair agent to \texttt{docfields.py}, where a revised split preserves the complete type. The terminal patch passes the frozen parenthesized-type test and fresh official verification, including a bracketed-type case. It retains the unrelated initial edit and is therefore successful but nonminimal.}
\label{app:trajectory-sphinx-9230}
\par\endgroup

\paragraph{Shared pattern.}
The two cases share a pattern that aggregate resolved rates do not show: useful generated tests identify a causal failure boundary rather than merely reporting that a patch is wrong. In Case~1, the recursion trace isolates a connector choice inside an otherwise plausible parity repair. In Case~2, the malformed rendered string redirects the agent from an unrelated subsystem to the parser on the failing path. Reusing the same frozen Test patch makes the before--after comparison executable, while the fresh official evaluator retains final authority. The protocol caveat in Case~1 and the nonminimal terminal patch in Case~2 also show why these examples are qualitative evidence rather than correctness certificates.

\section{Harness}
\label{app:formal-specifications}

\subsection{Fail-Closed Harness Semantics}
\label{app:harness-semantics}

This appendix gives a lower-level formal description of the same execution and submission protocol used in the main text. A Test bundle is $b=(\Delta_b,c_b,\kappa_b)$, where $\Delta_b$ is \texttt{test.patch}, $c_b$ is \texttt{test\_command}, and $\kappa_b$ is \texttt{test\_contract.json}. The task $x=(d,R_B)$ remains fixed throughout an episode. An immutable execution manifest $m$ records the repository commit, environment image, harness version, and reset receipt. Let
\begin{equation}
  r(b,R;m)\in\{0,1\}
  \label{eq:valid-run}
\end{equation}
indicate whether the artifacts validate, the required patches apply, the declared nodes bind, and execution completes without an operational error. Here $R$ is the source repository state before applying the Test patch $\Delta_b$. Conditional on $r(b,R;m)=1$, the harness returns a behavioral verdict $z$ and bounded output $o$,
\begin{equation}
  E(b,R;m)=(z,o),
  \qquad z\in\{\pass,\fail\},
  \label{eq:execution}
\end{equation}
which instantiates the main-text feedback interface on a candidate source patch as
\begin{equation}
  \mathcal{E}(b,p)=E(b,R_B\oplus p;m),
  \qquad r(b,R_B\oplus p;m)=1.
  \label{eq:verdict}
\end{equation}
\paragraph{Every candidate is a complete Base-relative source patch.}
The operator $\oplus$ denotes applying a source patch to the Base checkout. Each $p_t$ contains the complete cumulative source changes relative to the immutable $R_B$, not merely the changes since $p_{t-1}$. If an earlier round changes source region $A$ and a later round additionally changes region $B$, the later patch contains the retained changes to both $A$ and $B$ relative to Base, including any subsequent edits or reversions. CHECK and final submission use this same complete representation, including intended new source files and staged changes. The verifier resets to $R_B$ before applying $p_t$ and the fixed Test bundle; it does not reconstruct a candidate by chaining incremental Repair patches.

 When $r(b,R;m)=0$, the harness returns a bounded operational diagnostic rather than a behavioral verdict; it cannot establish local acceptance. Thus operational invalidity is tracked by $r$, not by adding a third value to the main-text $z$. When the protocol fixes the manifest, we omit $m$ from these expressions.

\paragraph{Behavioral failure is distinct from operational invalidity.}
For a valid run, the harness returns \pass{} when all bound test nodes pass and \fail{} when the completed test run reports a behavioral failure. A failed assertion or an exception caused by the exercised source behavior can supply Repair feedback; for example, the source-induced \texttt{RecursionError} in \cref{app:trajectory-django-16901} exposes a behavioral defect. By contrast, missing environment dependencies or a verification-infrastructure failure that prevents a valid test execution supplies only an operational diagnostic, not evidence that the patch failed the behavioral requirement. Classification depends on the cause and execution-validity evidence, not on a nonzero exit code or the test runner's error count alone. The contract $\kappa_b$ is structurally validated metadata, not an independent semantic judge. A schema error, patch-application failure, missing node, timeout, or infrastructure error sets $r(b,R;m)=0$ and cannot establish either verdict. The runtime envelope's \texttt{pass} denotes a valid passing run; \texttt{inconclusive} denotes an operationally invalid run. A \texttt{hard\_fail} is a behavioral \fail{} only when execution is valid; harness-blocked failures instead have $r=0$, as recorded by the failure-kind and validity diagnostics. Operational diagnostics never grant extra revision budget. Source-patch application failures use the recovery message in \cref{lst:selfrepair-apply-failure}; any continued repair remains within the same bounded episode.

For a fixed Test bundle, local acceptance means
\begin{equation}
  \operatorname{Accept}_{b}(p;m)
  =\begin{cases}
    \mathbb{1}\!\left[\operatorname{proj}_{z}E(b,R_B\oplus p;m)=\pass\right],
      &r(b,R_B\oplus p;m)=1,\\
    0,&r(b,R_B\oplus p;m)=0.
  \end{cases}
  \label{eq:verifier}
\end{equation}
\paragraph{A valid pass or budget exhaustion ends repair.}
As in \cref{sec:test-to-improve-loop}, local acceptance and official correctness are distinct. A valid local pass immediately terminates the Repair episode and forwards the passing patch without an additional agent turn or explicit submission action. The agent may also submit its current patch before a local pass. If no valid pass occurs within five revision rounds, the controller terminates the episode and force-submits the latest candidate, rather than reverting to an earlier patch. Within a feedback-guided episode, the forwarding rule is
\begin{equation}
  \begin{aligned}
    \operatorname{Forward}_{b}(p_t)
      &=\mathbb{1}\!\left[
        \begin{gathered}
          \operatorname{Accept}_{b}(p_t;m)=1
          \ \text{or explicit final submission}\\
          \text{or controller force-submission at budget exhaustion}
        \end{gathered}
      \right],\\
    \operatorname{Correct}_{x}(p_t)&=V_x^\star(p_t).
  \end{aligned}
  \label{eq:grounded-submit}
\end{equation}
A valid pass is sufficient for forwarding but does not replace the full official evaluation. A valid failure permits revision within budget or explicit submission of the current patch. An operationally invalid check is neither a pass nor a behavioral failure and does not grant additional rounds. If the final check is operationally invalid at budget exhaustion, the latest patch is still submitted; during training, the affected trajectory receives zero advantage as specified in \cref{app:repair-training-config}. If no Test bundle qualifies on Base, feedback-guided Repair is skipped and $p_0$ is forwarded for official evaluation without removing the instance from the denominator. The offline Test audit uses independently reset Base and Gold states. Making the manifests explicit, the same Base-to-Gold indicator as in \cref{eq:b2g-gate} is
\begin{equation}
  Q_x(b;m_B,m_G)=
  \begin{cases}
    \mathbb{1}[z_B=\fail]\,\mathbb{1}[z_G=\pass],
      &r(b,R_B;m_B)=r(b,R_G;m_G)=1,\\
    0,&\text{otherwise},
  \end{cases}
  \label{eq:base-gold-formal}
\end{equation}
Here $z_B=\operatorname{proj}_{z}E(b,R_B;m_B)$ and $z_G=\operatorname{proj}_{z}E(b,R_G;m_G)$ are defined only for valid executions. With fixed manifests, $Q_x(b;m_B,m_G)$ is abbreviated to $Q_x(b)=B_x(b)G_x(b)$. Operationally invalid Base or Gold runs count as zero for this evaluation metric; operational errors during candidate discrimination instead mask the affected training trajectory with zero advantage, as specified in \cref{sec:learn-to-test-obj}.

\subsection{Harness Updates Require Matched Evidence}
\label{app:harness-evolution}

\begin{figure}[!htbp]
  \centering
  \includegraphics[width=0.98\linewidth]{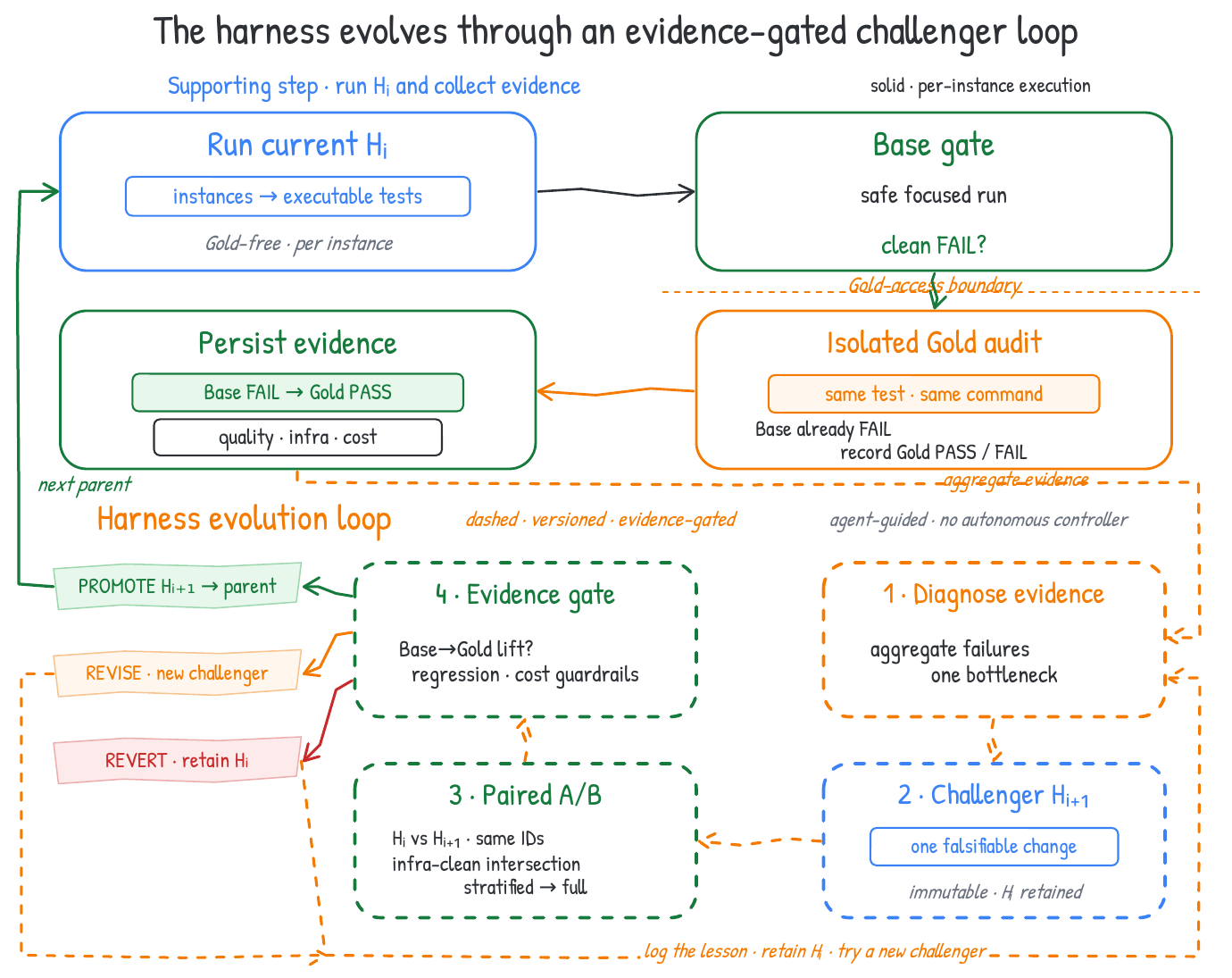}
  \caption{\textbf{Proposed evidence-gated harness maintenance, separate from the evaluated runtime.} The solid path records per-instance generation, Base qualification, and an isolated Gold audit. The dashed loop uses aggregated evidence to propose a single-change challenger and compare it with the retained parent on matched instances. The release owner promotes the challenger only if all predeclared quality, regression, infrastructure, and cost gates pass. No autonomous update or measured maintenance gain is claimed.}
  \label{fig:harness-evolution}
\end{figure}

We call the maintenance specification in \cref{fig:harness-evolution}
\emph{evidence-gated harness evolution}. Iterative code-and-test systems use
execution outcomes to propose subsequent candidates, and Meta-Harness extends
this pattern to executable harness code by letting a proposer inspect prior
code, scores, and execution traces~\citep{ravi2025llmloop,wang2025cure,jin2025reveal,wang2026codea1,lee2026metaharness}.
This proposed extension borrows only this candidate-generation pattern. It replaces
open-ended search and automatic candidate selection with a matched,
human-controlled maintenance trial for the versioned gentest harness, separate
from the evaluated \method{} runtime.

\paragraph{Each challenger isolates one declared change.}
Let $H_i$ be the retained parent and $H_{i+1}$ an immutable challenger with one
declared change and hypothesis. Under either version, the solid path generates
each Test patch without Gold, requires a clean Base failure, and then performs
an isolated offline Gold audit without changing \texttt{test.patch},
\texttt{test\_command}, or \texttt{test\_contract.json}. The resulting record
contains the versioned harness and prompt digests, instance and environment
manifests, patch-application receipts, Base and Gold verdicts, bounded logs,
cost, and infrastructure status. An agent may inspect the accumulated records,
select one recurring failure mode, and propose $H_{i+1}$; it cannot initiate the
trial or promote the challenger.

\paragraph{Paired trials predeclare IDs, metrics, and budgets.}
Before executing either version, the release owner freezes the paired instance
IDs and strata; the primary quality metric, direction, and minimum effect
threshold; the regression set; the model, sampling, retry, timeout, and seed
policy; the cost budget; and the treatment of infrastructure errors. The two
versions then run on the same IDs under independently recreated repository
state. A quality pair is admissible only if both versions have verified patch
application, complete logs, and no infrastructure error. The primary quality
comparison uses only these jointly admissible pairs, whereas infrastructure
error and cost rates retain all predeclared paired IDs in their denominators.
No post hoc removal, replacement, or relabeling of an ID changes a gate.

\paragraph{Promotion requires every declared gate to pass.}
The release owner promotes $H_{i+1}$ only if every predeclared gate passes: the
paired quality improvement reaches its threshold, no regression gate fails,
the infrastructure policy is satisfied, and the cost budget is met. Otherwise
$H_i$ remains the released version. A revision becomes a new immutable
challenger and a reversion restores $H_i$; in both cases the decision record
retains the failed hypothesis and all trial artifacts. Thus an improvement on
the admissible quality subset cannot conceal an increase in missingness,
infrastructure failures, or cost.

\paragraph{Harness evolution remains outside the runtime.}
This paper reports no matched maintenance trial and therefore no effect of the
policy. The dashed path lies outside the \method{} runtime: no autonomous
controller modifies or promotes the gentest harness. Gold is used only in the
isolated offline audit and never enters a Test-agent episode, candidate
artifact, or online Base-gate feedback. The full official evaluator remains
the authority for Repair-patch correctness.

\subsection{Prompt and Controller Records}
\label{app:harness-prompts}

The remaining records specify the fixed messages and dynamic feedback envelope used by the generated-test patch harness. Round~0 denotes the initial solve that produces the source Repair patch before generated-test repair. A complete seeded conversation also contains the issue message, prior assistant and tool turns, shell observations, the Bash tool schema, and candidate-check receipts. We omit inherited generic adapter and tool definitions.

The listings omit YAML indentation and Python string delimiters that the parsers remove before inference. Double-braced fields are Jinja placeholders for initial messages; single-braced fields use Python \texttt{format} for follow-ups. The harness fills \texttt{task}, \texttt{step\_limit}, \texttt{error}, and bounded \texttt{feedback} at runtime. Tool schemas, repository contents, observations, and dynamic validator diagnostics enter separately and are not reproduced below.

\subsubsection{Test-Agent Prompt and Submission Protocol}

The harness sends \cref{lst:test-system} once as the system message and \cref{lst:test-instance} once as the instance message. The model receives the public issue and a buggy Base checkout. Gold, the reference Repair patch, and official tests remain outside its context. A malformed tool response triggers \cref{lst:test-format-error}.

\begin{promptbox}{Test-agent system message (verbatim).}{lst:test-system}
You are an experienced repository maintainer writing one precise regression test for the issue.
Correctly identify the smallest public behavior contract before treating a Base failure as useful.
You have one standard bash tool; all inspection, editing, testing, and submission use it.
\end{promptbox}

\begin{promptbox}{Test-agent instance message (verbatim template).}{lst:test-instance}
You are in a real shell at /testbed, checked out at the BUGGY commit: the bug is NOT fixed.
A correct regression test should cleanly FAIL here for the issue behavior, and would pass once
a correct fix is applied. You do not have that fix, its patch, or oracle tests.

<original_issue>
{{task}}
</original_issue>

The prepared repository environment is active and the package is importable. Do not create a
virtualenv, reinstall, or run pip. Infer the repository-native test entrypoint and selector from
current configuration, CI, and adjacent tests. Do not create a standalone reproduction script,
custom settings module, or custom runner when an adjacent repository test can exercise the issue.

You have at most {{ step_limit }} model turns. A clean Base failure is necessary but not sufficient:
it proves only that the buggy checkout disagrees with your assertion, not that the assertion is the
behavior required by the issue. Before editing, determine the public behavior contract. Inspect the
adjacent test file, its fixtures, parameter tables, and nearby boundary cases so your test uses the
same supported workflow. Consider at least one plausible alternative interpretation of the issue.

Gold-free boundary:
- Never inspect git history, refs, branches, tags, remotes, commits, or `.git` internals. Do not run
  `git log`, `git show`, `git branch`, `git cat-file`, or any diff against a commit/ref.
- For git, use only `git status`, `git add -AN`, and plain working-tree `git diff`.
- Derive expected behavior only from the issue text, public docs, current buggy code, and current
  tests. Do not search for or reconstruct the historical fix or oracle test.

Test-quality contract:
1. Exercise a public API, CLI, rendered output, migration operation, or established repository
   workflow. Do not assert private/internal state merely because it is easy to inspect. If the
   issue is about rendering or integration behavior, test that observable result rather than an
   internal flag or helper.
2. Assert the CORRECTED behavior, never the buggy symptom. Pin an exact value, message, count, or
   ordering only when the issue, docs, or an existing public test explicitly fixes that literal.
   Otherwise use the weakest observable property that the issue guarantees and buggy Base violates.
3. Add one coherent regression-test group for this behavior. It may contain one or more new
   `test_*` functions or methods and may modify the fixture, parameter, helper, or data files they
   require. If the repository already expresses cases through a parameter/data table, extending
   that table and targeting its existing exact test node is allowed. Do not add unrelated coverage.
4. The one-line command in /tmp/test_command must explicitly select every test node declared for
   this behavior. A file-, class-, directory-, or suite-level command is rejected even if it fails.

Before submission, write valid JSON to /tmp/test_contract.json with exactly these required fields:
  - "entrypoint": public API/CLI/workflow exercised;
  - "trigger": issue-backed input and setup that exposes the bug;
  - "expected_output": minimal corrected observable behavior asserted;
  - "issue_evidence": nonempty list of issue/doc/current-test evidence supporting the expectation;
  - "alternative_hypotheses": nonempty list of plausible interpretations you considered and rejected;
  - "non_assertions": nonempty list of tempting literals/internal details intentionally not pinned;
  - "test_nodes": a nonempty list containing every exact test_* name exercised by this change.
The harness validates this structure, requires every newly added test_* to be declared, and binds
every declared node to /tmp/test_command. Existing nodes are allowed for fixture/parameter/data edits.

Protocol:
1. Inspect only enough implementation to confirm a public call signature or observable behavior,
   plus the adjacent tests/fixtures/parameters needed for a repository-native test.
2. Edit files directly in /testbed with bash. Run the exact focused command for the declared nodes. If it
   fails for the intended issue behavior, review every assertion against the behavior contract;
   revise over-pinned, private-state, or unsupported expectations before submitting.
3. In one preparation bash call, write /tmp/test_contract.json and exactly one command line to
   /tmp/test_command, then create and inspect the patch outside the repository:

     cd /testbed && git add -AN && git diff --no-color > /tmp/test.patch

   /tmp/test.patch must be nonempty and contain only the intended regression-test changes.
4. Submit with exactly one final bash call:

   echo COMPLETE_TASK_AND_SUBMIT_FINAL_OUTPUT && cat /tmp/test.patch

   The harness applies the patch on an isolated clean buggy Base and runs only the bound test nodes.
   If rejected, fix the reported contract, selector, collection, or behavior problem and resubmit.

Nothing is submitted by ordinary prose. Every response must contain a bash tool call.
\end{promptbox}

\begin{promptbox}{Test-agent response-format repair message (verbatim template).}{lst:test-format-error}
Tool call error:

<error>
{{error}}
</error>

Every response must contain at least one standard bash tool call. Do not emit custom tools.
\end{promptbox}

\subsubsection{Repair-Agent Prompt and Controller Protocol}

With the isolated generated-test gate, the repair harness preserves the Round-0 source patch and evaluates candidates in a separate workspace. The Repair agent receives the issue, gate metadata, test filename, runner command, and a bounded output tail; the generated Test patch remains hidden, although tracebacks can reveal paths and assertion lines. The generation harness persists the Test patch, execution command, and behavior contract as one bundle. The gate uses the patch and command for execution, while the contract supports qualification and audit without entering the Repair-agent context.

The map's \texttt{source\_trajectory} and \texttt{source\_exit\_status} fields describe generated-test construction rather than repair seed history. For the post-trained Repair condition, the Round-0 trajectories come from the Test-to-Improve-trained Qwen-3.5-35B-A3B Repair agent. Evaluation uses three repair rollouts per source and 128 workers, as specified in \cref{tab:evaluation-hparams}. Round~0 uses 200 turns, followed by at most five 40-turn Repair revision rounds, with a forced candidate check at the end of each round and one fixed generated Test bundle per issue, shared across the three Repair rollouts. If the verifier has not passed after the fifth revision, the controller force-submits the latest candidate.

In provided-patch mode, only \texttt{-{}-seed-trajectory} adds trajectory context. The orchestrator converts the supplied source-row messages directly. A direct caller that passes \texttt{source\_messages=None} instead loads the trajectory from \texttt{-{}-source-run}.

\paragraph{Patch export preserves the Base-relative invariant.}
The CHECK and FINAL-SUBMIT commands below export the complete candidate defined in \cref{app:harness-semantics}. Before using a working-tree \texttt{git diff}, the tracked index entries must remain at Base and all intended new source files must be registered with intent-to-add. Any staged source edits must remain included in the final Base-relative export rather than silently changing the comparison point. The exported patch must reproduce the current candidate when applied to a fresh Base checkout; an incremental diff against the preceding round is not a valid substitute. All exports remain source-only and exclude the fixed Test bundle.

\paragraph{Seeded conversations preserve source context.}
For these solve trajectories, seed conversion retains system, user, and assistant text plus paired tool calls and shell observations. It drops exit rows, reasoning items, and unpaired tool items. The harness then appends \cref{lst:selfrepair-initial} as the next user message without inserting \cref{lst:selfrepair-system}. The post-trained Repair agent's Round-0 trajectory supplies the retained system message in \cref{lst:selfrepair-seeded-system}. The fallback path uses \cref{lst:selfrepair-system} only when no seed messages exist; the patch-application-failure branch uses the same fallback with \cref{lst:selfrepair-apply-failure}. The harness caps generated-test feedback at $6{,}000$ characters and patch-application diagnostics at $12{,}000$ characters.

\begin{promptbox}{System message retained by a seeded \texttt{swerebench} Round-0 conversation (verbatim).}{lst:selfrepair-seeded-system}
You are a helpful assistant that can interact with a computer shell to solve programming tasks.
\end{promptbox}

\begin{promptbox}{Self-repair fallback system message for an empty seed (verbatim).}{lst:selfrepair-system}
You are a careful software engineer that interacts with a computer shell to solve programming tasks.
\end{promptbox}

\begin{promptbox}{Self-repair initial user message after a non-passing, non-blocking initial generated-test gate (template).}{lst:selfrepair-initial}
You are fixing a bug. A previous patch attempt is ALREADY APPLIED in the repository workspace,
but the hidden regression-test gate has not produced a valid pass. You CANNOT see the test file;
you receive the execution feedback below.
Your job: edit the SOURCE code in /testbed so the described behavior is correct. Do NOT edit or
create test files.

<original_issue>
{{task}}
</original_issue>

<failing_test_output>
{{feedback}}
</failing_test_output>

A valid behavioral failure is useful evidence, but the generated test may be incomplete,
over-specific, or wrong. An operational error is not evidence that the source patch violates
the test requirement. Judge valid test feedback together with the original issue, repository
behavior, and checks you can run in the repository workspace.

Steps:
1. `cd /testbed && git diff` to see the already-applied previous attempt.
2. Distinguish behavioral failure from operational error. Infer the intended behavior from
   the original issue and any valid behavioral feedback.
3. Make the minimal SOURCE edit(s) needed. Only non-test files.
4. Re-derive your own quick check if helpful, then explicitly choose one action below.

Export the complete cumulative SOURCE patch relative to Base, including intended new files,
not just edits since the previous round. Preserve this rule for both CHECK and FINAL-SUBMIT.

To CHECK the current candidate against the hidden generated test, use EXACTLY two separate
commands (must exit 0):
  cd /testbed && git diff -- <non-test source files> > /tmp/fix.patch
  echo COMPLETE_TASK_AND_CHECK_CANDIDATE_PATCH && cat /tmp/fix.patch

The check returns environment feedback. A valid pass immediately ends repair; the controller
submits the passing patch without another agent action. A valid failure allows another revision
while budget remains and does not by itself prove the source patch is wrong. An operational
error is reported as inconclusive and does not add revision budget.
If the test has not passed after the fifth revision, the controller force-submits the latest
candidate and ends repair without requiring another agent decision.

To FINAL-SUBMIT the current patch because it is your best solution, even if the hidden generated
test is still failing or suspect, use EXACTLY two separate commands (must exit 0):
  cd /testbed && git diff -- <non-test source files> > /tmp/fix.patch
  echo COMPLETE_TASK_AND_SUBMIT_FINAL_OUTPUT && cat /tmp/fix.patch

FINAL-SUBMIT ends repair. The harness records the latest gate result separately; your decision
to submit does not claim that the generated test passed.

\end{promptbox}

\paragraph{Runtime feedback is structured and bounded.}
The initial and failure-follow-up messages embed the field order shown in \cref{lst:selfrepair-feedback}. The harness derives these values from its execution record, truncates each command and output field to its configured bound, and reserves the remaining character budget for the raw output tail. Angle-bracketed values below describe runtime data and are not literal prompt text.

\begin{promptbox}{Self-repair runtime feedback envelope (field order; values schematic).}{lst:selfrepair-feedback}
test_execution_contract:
  language: python
  working_directory: /testbed
  test_source_visibility: hidden_verifier_only
  per_test:
    {"index": <index>, "runner_command": "<command>", "test_filename": "<path>"}
generated_test_gate:
  gate_status: <pass | hard_fail | inconclusive>
  gate_reason: <failure kind>
  passed: <True | False>
  n_tests: <count>
  failed_test_names: <JSON list>
  per_test:
    {"blocked_by_harness": <bool>,
     "command": "<activated command, present for a custom command>",
     "command_sanitized": <bool>,
     "counts": <object>, "failed_test_names": <list>,
     "failure_kind": <kind>, "filename": "<path>",
     "gate_status": <status>, "inconclusive_infra": <bool>,
     "index": <index>, "rc": <return code>,
     "removed_command_flags": <list>, "timeout_detected": <bool>}

<raw_test_output_tail>
<bounded stdout and stderr tail>
</raw_test_output_tail>
\end{promptbox}

\paragraph{Gate outcomes determine the next controller message.}
Under the canonical duplicate-continuation setting, an unchanged candidate CHECK selects \cref{lst:selfrepair-duplicate} without rerunning the gate or replenishing the turn budget. A changed candidate CHECK immediately ends repair on a valid pass, with the terminal notice in \cref{lst:selfrepair-pass-followup}. A non-passing check with revision budget remaining returns the non-passing follow-up, distinguishing valid behavioral failure from operational invalidity in the feedback envelope. Forty turns without a candidate check force one, and each episode permits at most five feedback-guided revision rounds after Round~0. A valid passing candidate is automatically submitted by the controller; no further source revision or explicit agent submission is required. The agent may also explicitly submit the current patch before the budget is exhausted even if the test still fails. If the verifier has not passed after the fifth revision, the controller force-submits the latest candidate and terminates repair without another agent decision; \cref{lst:selfrepair-final-decision} records the corresponding terminal notice. The controller does not offer a separate keep-original action. This lifecycle uses local acceptance for stopping and forwarding, while leaving official correctness to the full evaluator, as formalized in \cref{eq:grounded-submit}.

\begin{promptbox}{Self-repair follow-up after a non-passing candidate check with revision budget remaining (template).}{lst:selfrepair-fail-followup}
Your latest source patch was tested against the hidden generated regression test, but the gate
has not produced a valid pass. Distinguish behavioral failure from an operationally inconclusive
execution; an environment or infrastructure error does not establish a source-code defect.
Treat the output as environment feedback, not an absolute veto. Compare it
with the original issue and repository behavior. Then either make another minimal source-only
edit and CHECK the candidate again, or FINAL-SUBMIT the current patch if the generated test is
wrong or over-specific and the source patch is already the best solution.

<updated_failing_test_output>
{feedback}
</updated_failing_test_output>

Do not edit or create tests. Export the full cumulative SOURCE patch relative to Base,
including intended new source files, for either action.

CHECK candidate:
  cd /testbed && git diff -- <non-test source files> > /tmp/fix.patch
  echo COMPLETE_TASK_AND_CHECK_CANDIDATE_PATCH && cat /tmp/fix.patch

FINAL-SUBMIT current patch regardless of the gate result:
  cd /testbed && git diff -- <non-test source files> > /tmp/fix.patch
  echo COMPLETE_TASK_AND_SUBMIT_FINAL_OUTPUT && cat /tmp/fix.patch

\end{promptbox}

\begin{promptbox}{Self-repair terminal notice after a valid passing candidate check (template).}{lst:selfrepair-pass-followup}
The hidden generated-test CHECK passed in a valid execution.
The controller has submitted the passing candidate as the final source patch.
Repair is complete; no further edits, checks, or submission decisions are accepted.
The full official evaluator determines whether this patch resolves the task.
\end{promptbox}

\begin{promptbox}{Self-repair terminal notice after exhausting the revision budget (template).}{lst:selfrepair-final-decision}
The generated test has not produced a valid pass after the fifth feedback-guided revision.
The controller has force-submitted the latest candidate as the final source patch.
Repair is complete; no further edits, checks, or submission decisions are accepted.
The full official evaluator determines whether this patch resolves the task.
\end{promptbox}

\begin{promptbox}{Self-repair follow-up after an unchanged candidate check (verbatim).}{lst:selfrepair-duplicate}
The latest candidate CHECK did not make a new source-code change, so the hidden generated test was
not rerun and this did not consume a gate check. Inspect the current diff and feedback. Either make
a concrete non-test source edit before checking again, or explicitly FINAL-SUBMIT the unchanged
patch if it is already the best solution and the generated test is wrong or over-specific.
\end{promptbox}

\paragraph{Source-patch recovery preserves the qualification boundary.}
For a qualified Test bundle, an empty Round-0 source patch leaves the repository at Base and must reproduce the clean Base failure under the same deterministic execution conditions. An empty-patch pass is therefore not a normal entry state of the qualified-bundle protocol; it indicates inconsistent execution evidence rather than a reason to bypass Base qualification. When a provided source patch cannot be applied, the harness instead uses \cref{lst:selfrepair-apply-failure} to request a valid source-only patch while preserving the fixed Test bundle and the existing episode budget.

\begin{promptbox}{Self-repair recovery message after source-patch application failure (verbatim template).}{lst:selfrepair-apply-failure}
You are fixing a bug. A previous patch attempt could NOT be applied by the harness, so /testbed
is currently at the clean base checkout. You CANNOT see the hidden regression tests yet.

Your job in this round: create a valid minimal SOURCE patch in /testbed that addresses the
original issue. Use the failed patch/apply diagnostics only as hints. Do NOT edit or create test
files.

<original_issue>
{{task}}
</original_issue>

<failed_patch_apply_diagnostics>
{{feedback}}
</failed_patch_apply_diagnostics>

Steps:
1. `cd /testbed && git status && git diff` to confirm the clean starting point.
2. Read the issue and the patch-apply diagnostics. Infer the intended source change.
3. Make the minimal SOURCE edit(s). Only non-test files.
4. Submit a syntactically valid unified diff.

Submit with EXACTLY two separate commands (must exit 0):
  cd /testbed && git diff -- <non-test source files> > /tmp/fix.patch
  echo COMPLETE_TASK_AND_SUBMIT_FINAL_OUTPUT && cat /tmp/fix.patch
\end{promptbox}

\end{document}